\documentclass{article}
\usepackage{iclr2027_conference,times}

\usepackage{amsmath,amsfonts,bm}

\def\eqref#1{equation~\ref{#1}}

\def\1{\bm{1}}

\DeclareMathAlphabet{\mathsfit}{\encodingdefault}{\sfdefault}{m}{sl}
\SetMathAlphabet{\mathsfit}{bold}{\encodingdefault}{\sfdefault}{bx}{n}

\DeclareMathOperator*{\argmax}{arg\,max}

\usepackage{hyperref}
\usepackage{url}
\usepackage{graphicx}
\usepackage{booktabs}
\usepackage{amsmath}
\usepackage{xcolor}
\usepackage{cleveref}
\creflabelformat{equation}{#2#1#3}
\usepackage{wrapfig}
\usepackage{algorithm,algpseudocode}
\usepackage{enumitem}
\usepackage{booktabs}
\usepackage{multirow}
\usepackage{xparse}

\usepackage{times,amsmath,amssymb,graphicx,tikz}
\usetikzlibrary{arrows.meta,positioning,calc,fit,backgrounds,
                decorations.pathreplacing,shapes.geometric,patterns}

\usetikzlibrary{arrows.meta, positioning, calc, decorations.pathmorphing}

\definecolor{rot}{RGB}{200,60,55}      
\definecolor{blau}{RGB}{45,95,175}     
\definecolor{grau}{RGB}{120,120,120}   
\definecolor{gruen}{RGB}{35,130,90}    
\definecolor{lgrau}{RGB}{242,242,244}  
\definecolor{mgrau}{RGB}{205,205,210}  
\colorlet{zc1}{blau}
\definecolor{zc2}{RGB}{30,140,140}
\definecolor{zc3}{RGB}{220,120,30}
\definecolor{zc4}{RGB}{130,70,160}

\tikzset{
  ax/.style   = {->,>={Stealth[length=2.2mm,width=1.7mm]},line width=.5pt,grau},
  axb/.style  = {->,>={Stealth[length=3.0mm,width=2.4mm]},line width=1.0pt},
  panel/.style= {rounded corners=2pt,fill=lgrau,draw=mgrau,line width=.4pt},
  box/.style  = {rounded corners=1.5pt,draw=grau,fill=white,line width=.5pt,
                 inner sep=3.5pt,font=\small,align=center},
  hd/.style   = {font=\normalsize\bfseries,align=center},
  lb/.style   = {font=\small,align=center},
  tny/.style  = {font=\scriptsize,align=center,text=grau},
}

\newcommand{\dtile}[2][1.55cm]{\includegraphics[width=#1]{figures/tiles/#2.pdf}}

\makeatletter
\NewDocumentCommand{\LeftComment}{s m}{%
  \Statex \IfBooleanF{#1}{\hspace*{\ALG@thistlm}}\(\triangleright\) #2}
\makeatother

\title{Occupancy Modeling via Drifting: \\ One-Step Generation Enables \\ Direct Policy Extraction}

 \author{%
    \vspace{-0.1cm}
    \parbox[t]{\dimexpr\linewidth-2\tabcolsep}{\centering\normalfont
        \textbf{Robin Uhrich}$^{1}$ \hspace{0.25cm}
        \textbf{Marco Bagatella}$^{*,2,3}$ \hspace{0.25cm}
        \textbf{N\'uria Armengol Urp\'i}$^{*,2,3}$ \\[0.8ex]
        \makebox[\linewidth][c]{%
            \textbf{Pavel Kolev}$^{1}$ \hspace{0.25cm}
            \textbf{Georg Martius}$^{1}$%
        } \\[1.2ex]
        $^{1}$University of T\"ubingen \hspace{0.25cm} $^{2}$ETH Z\"urich \hspace{0.25cm}
        $^{3}$Max Planck Institute for Intelligent Systems \\
        * Equal supervision
    \vspace{-0.1cm}
  }
}

\iclrfinalcopy 
\begin{document}

\maketitle

\begin{abstract}

The growing interest in behavior foundation models, combined with recent advancements in generative modeling, has led to an appealing recipe for reinforcement learning from unlabeled data: pre-train a generative model to model occupancies, i.e., discounted state visitations, on reward-free trajectories, such that a policy can be easily extracted once a small reward-labeled dataset becomes available.
We decompose this recipe along three axes: (i) the degree to which the multi-modal structure of occupancies must be captured, (ii) the class of generative model, and (iii) the method used for policy extraction.
Along each axis, we consider existing choices, propose simpler alternatives, and validate each decision experimentally.
This analysis culminates in a novel, streamlined method that leverages drifting, a modern one-step generative sampler, to model the occupancy.
We show that this enables policy extraction via direct differentiation through the occupancy model, and bypasses the need for critic distillation. Overall, we observe that this nimble setup leads to improvements in performance on standard benchmarks. \looseness -1
\end{abstract}

\section{Introduction}
Reinforcement learning (RL) requires rewards, yet most available interaction data comes without them.
In the absence of rewards, instead of learning how desirable each behavior is, it is convenient to learn to predict the states each behavior visits~\citep{touati2023,farebrother2025,zheng2025}.
Then, once a reward is revealed, this knowledge can be readily leveraged to extract a good policy~\citep{zheng2025}.
This is the intuition motivating occupancy models, which predict the discounted distribution over future states~\citep{janner2020, blier2021} that will be visited when starting from a given state and following a policy.
If this is estimated correctly, computing values for any given reward becomes trivial (see \Cref{sec:background}) and computationally efficient.
In turn, this facilitates downstream policy selection or extraction \citep{schramm2024, farebrother2025, zheng2025}.

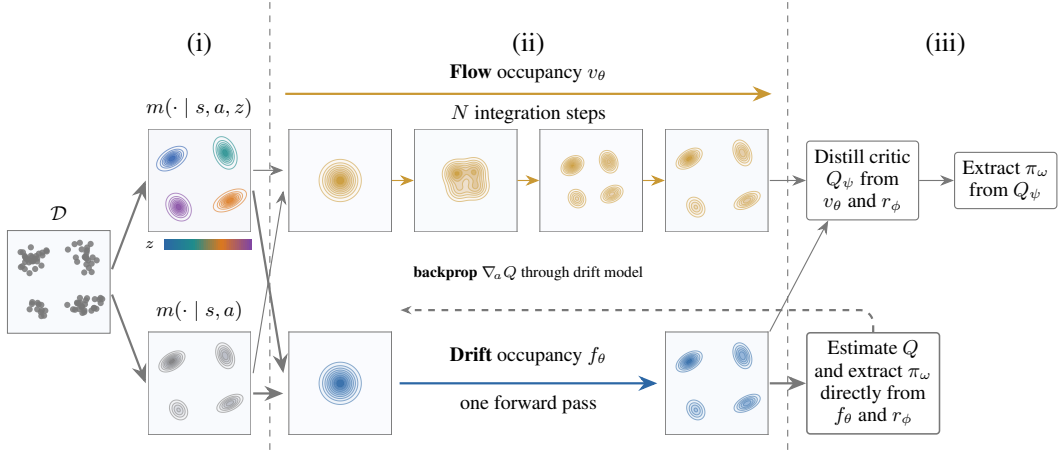
\begin{figure}[h]
    \centering
    \resizebox{\linewidth}{!}{
\begin{tikzpicture}[x=1cm,y=1cm,
  ax/.style  = {->,>={Stealth[length=2.2mm,width=1.7mm]},line width=.5pt,grau},
  axb/.style = {->,>={Stealth[length=3.0mm,width=2.4mm]},line width=1.0pt},
  box/.style = {rounded corners=1.5pt,draw=grau,fill=white,line width=.5pt,
                inner sep=3.5pt,font=\small,align=center},
  lb/.style  = {font=\small,align=center},
  tny/.style = {font=\scriptsize,align=center,text=black}]
\definecolor{blau}{RGB}{44,103,165}     
\definecolor{gelb}{RGB}{201,152,51}     
\definecolor{grau}{RGB}{120,120,120}    
\definecolor{tilebg}{RGB}{248,250,253}  
\colorlet{zc1}{blau}                    
\definecolor{zc2}{RGB}{30,140,140}
\definecolor{zc3}{RGB}{220,120,30}
\definecolor{zc4}{RGB}{130,70,160}

\def\ts{1.62cm}                                  
\def\xa{1.20}\def\xb{3.18}\def\xc{5.16}\def\xd{7.14}   
\def\yF{3.20}\def\yD{0.00}                       
\def\mx{-1.02}                                   
\def\axisIx{\mx}                                 
\def\axisIIx{{(\xa+\xd)/2}}
\def\axisIIIx{11.5}

\node[lb,anchor=south] at (-3.24,2.46) {$\mathcal{D}$};
\begin{scope}[shift={(-3.24,1.60)},x=\ts,y=\ts]
  \fill[tilebg] (-.5,-.5) rectangle (.5,.5);
  \foreach \x/\y in {
  -0.331/0.125, -0.169/0.159, -0.23/0.126, -0.194/0.275, -0.159/0.258,
  -0.234/0.226, -0.328/0.194, -0.355/0.153, -0.22/0.252, -0.349/0.288,
  -0.253/0.2, -0.246/0.206, -0.29/0.19, -0.121/0.307, -0.272/0.254,
  -0.117/0.297, -0.372/0.29, -0.369/0.144, -0.272/0.337, -0.285/0.085,
  -0.21/0.228, -0.309/0.222, -0.26/0.241, -0.331/0.252, -0.16/0.285,
  -0.32/0.229, -0.211/0.2, -0.327/0.241, -0.27/0.203, 0.253/0.244,
  0.166/0.238, 0.293/0.136, 0.276/0.335, 0.257/0.238, 0.369/0.21,
  0.33/0.238, 0.248/0.092, 0.274/0.307, 0.205/0.381, 0.32/0.178,
  0.179/0.196, 0.32/0.096, 0.286/0.21, 0.154/0.368, 0.31/0.37,
  0.223/0.317, 0.27/0.285, 0.271/0.227, 0.089/0.315, 0.265/0.251,
  0.269/0.17, -0.196/-0.219, -0.258/-0.197, -0.206/-0.204, -0.143/-0.329,
  -0.175/-0.295, -0.21/-0.267, -0.284/-0.267, -0.261/-0.215,
  -0.165/-0.277, -0.131/-0.28, -0.17/-0.224, -0.17/-0.333, 0.396/-0.172,
  0.159/-0.271, 0.227/-0.239, 0.255/-0.218, 0.083/-0.224, 0.423/-0.167,
  0.222/-0.246, 0.351/-0.168, 0.366/-0.262, 0.181/-0.209, 0.345/-0.253,
  0.248/-0.295, 0.193/-0.21, 0.263/-0.18, 0.341/-0.255, 0.206/-0.156,
  0.25/-0.278, 0.407/-0.156, 0.275/-0.159, 0.139/-0.267, 0.23/-0.302,
  0.21/-0.203, 0.318/-0.28, 0.178/-0.202, 0.408/-0.206, 0.197/-0.264}
    {\fill[grau,fill opacity=.85] (\x,\y) circle (0.03);}
  \draw[grau,line width=.4pt] (-.5,-.5) rectangle (.5,.5);
\end{scope}

\node[lb,anchor=south] at (\mx,4.06) {$m(\cdot\mid s,a, z)$};
\node[inner sep=0] (mz) at (\mx,3.20) {\dtile[\ts]{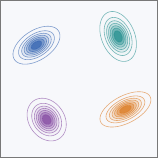}};
\foreach \i in {0,...,44} {
  \pgfmathsetmacro{\t}{\i/44*3}
  \pgfmathtruncatemacro{\sg}{min(floor(\t),2)}
  \pgfmathtruncatemacro{\ca}{\sg+1}
  \pgfmathtruncatemacro{\cb}{\sg+2}
  \pgfmathtruncatemacro{\pc}{round(100-(\t-\sg)*100)}
  \fill[zc\ca!\pc!zc\cb] ({-1.57+\i*1.36/45},2.11) rectangle ++({1.36/45+0.004},0.17);
}
\node[tny,anchor=east] at (-1.61,2.195) {$z$};
\node[lb,anchor=south] at (\mx,0.86) {$m(\cdot\mid s, a)$};
\node[inner sep=0] (mm) at (\mx,0.00) {\dtile[\ts]{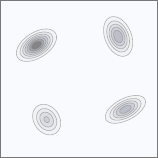}};

\node[inner sep=0] (f0) at (\xa,\yF) {\dtile[\ts]{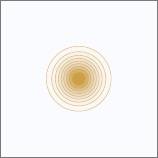}};
\node[inner sep=0] (f1) at (\xb,\yF) {\dtile[\ts]{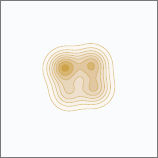}};
\node[inner sep=0] (f2) at (\xc,\yF) {\dtile[\ts]{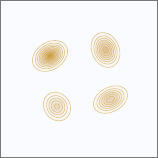}};
\node[inner sep=0] (f3) at (\xd,\yF) {\dtile[\ts]{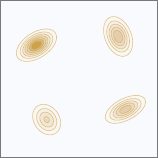}};
\node[inner sep=0] (d0) at (\xa,\yD) {\dtile[\ts]{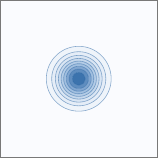}};
\node[inner sep=0] (d1) at (\xd,\yD) {\dtile[\ts]{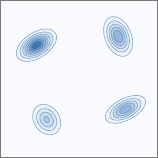}};

\draw[axb,gelb] ($(f0.north west)+(-0.05,0.55)$) -- ($(f3.north east)+(0.05,0.55)$);
\node[lb,anchor=south] at ({(\xa+\xd)/2},\yF+1.43) {\textbf{Flow} occupancy $v_\theta$};
\node[lb,anchor=north] at ({(\xa+\xd)/2},\yF+1.31) {$N$ integration steps};
\foreach \a/\b in {f0/f1, f1/f2, f2/f3} {\draw[ax,gelb] (\a) -- (\b);}

\draw[axb,blau] ($(d0.east)+(0.12,0)$) -- ($(d1.west)+(-0.12,0)$);
\node[lb,anchor=south] at ({(\xa+\xd)/2},\yD+0.16) {\textbf{Drift} occupancy $f_\theta$};
\node[lb,anchor=north] at ({(\xa+\xd)/2},\yD-0.06) {one forward pass};

\node[box,anchor=west] (ex1) at (8.55,\yF)
      {Distill critic \\$Q_\psi$ from \\$v_\theta$ and $r_\phi$};
\node[box,anchor=west] (ex1b)
      at ($(ex1.east)+(0.55,0)$) {Extract $\pi_\omega$\\ from $Q_\psi$};
\node[box,anchor=west,line width=.7pt,
      pattern color=grau!35] (ex2) at (8.55,\yD)
      {Estimate $Q$ \\ and extract $\pi_\omega$\\ directly from \\$f_\theta$ and $r_\phi$};

\draw[axb,grau] (-2.40,1.80) -- ($(mz.west)+(-0.03,0)$);
\draw[axb,grau] (-2.40,1.40) -- ($(mm.west)+(-0.03,0)$);
\foreach \t/\y in {mz/0.16, mm/-0.16} {
  \draw[ax]       ($(\t.east)+(0.03, 0.16)$) -- ($(f0.west)+(-0.08,\y)$);
  \draw[axb,grau] ($(\t.east)+(0.03,-0.16)$) -- ($(d0.west)+(-0.08,\y)$);
}
\draw[ax]       ($(f3.east)+(0.02,0)$) -- (ex1.west);
\draw[ax]       (ex1) -- (ex1b);
\draw[axb,grau] ($(d1.east)+(0.02,0)$) -- (ex2.west);
\draw[ax]       ($(d1.north east)+(0.02,0)$) -- ($(ex1.south west)+(0.35,0)$);
\coordinate (bpX) at (\xa+0.95,0);
\coordinate (bpMidY) at ($(ex2.north)!0.5!(0,{\yD+1.5})$);
\coordinate (bpEnd) at (bpMidY -| bpX);
\draw[axb,grau,dashed,rounded corners=6pt,-{Stealth[length=5pt]}]
  (ex2.north) -- (bpEnd -| ex2.north) -- (bpEnd);
\node[tny,anchor=south] at ({(\xb+\xc)/2},{\yD+1.5})
  {\textbf{backprop} $\nabla_{\!a}Q$ through drift model};

\node[tny,anchor=south] at (\axisIx,\yF+1.85)   {\large{(i)}};
\node[tny,anchor=south] at (\axisIIx,\yF+1.85)  {\large{(ii)}};
\node[tny,anchor=south] at (\axisIIIx-0.75,\yF+1.85) {\large{(iii)}};
\foreach \x in {0.09,8.26}
  {\draw[grau,dashed,line width=.5pt] (\x,-1.05) -- (\x,6.05);}

\end{tikzpicture}}
    \vspace{-4mm}
    \caption{A generative pipeline for unsupervised RL. Starting from an unlabeled dataset $\mathcal{D}$, the occupancy can be modeled in a conditional or unconditional fashion~(i). Independently of this choice, the occupancy can be learned through flow matching or drifting~(ii). In both cases, the policy can be extracted in a two-step process, but drifting also allows for direct policy extraction~(iii). We study all combinations of these choices, and recommend those marked by a bold arrow.}
    \label{fig:overview}
\end{figure}

Recent works have trained expressive generative models to model occupancies from data, largely through a temporal difference objective~\citep{farebrother2025, zheng2025}.
While remarkably expressive, they often involve complex, multi-step pipelines.
Intention-Conditioned Flow Occupancy Models (InFOM)~\citep{zheng2025} rely on a flow-based generative model of the occupancy, which requires several costly integration steps per sample.
Moreover, this choice of generative model complicates the policy extraction procedure.
Once a reward is revealed, a differentiable occupancy model would allow training an actor in a policy-gradient fashion by backpropagating through the occupancy to find actions that lead to high-reward states in the future.
As differentiation through the ODE is not straightforward, \citet{zheng2025} resort to a two-step process for policy extraction, by first distilling a critic from Monte Carlo value estimates, and then using deep policy gradients of the distilled critic to update the actor's parameters.
Finally, InFOM conditions the occupancy model on a latent intention inferred by a variational encoder, in order to perform generalized policy improvement (GPI)~\citep{barreto2017, farebrother2025} over behaviors present in the data.
While this is in principle a powerful policy extraction technique, in practice it relies on accurate inference of each mode of the data generation process.
Each additional component adds computational requirements, hyperparameters, and potential failure modes, while its importance in practical settings has so far been explored only partially.

In this work, \textbf{we revisit this generative pipeline for unsupervised RL}, and isolate three design choices that are central to occupancy modeling and policy extraction (see ~\Cref{fig:overview}):
\begin{enumerate}[label=(\roman*),leftmargin=*]
    \item \textit{Target.} The occupancy can be modeled in a conditional fashion, by inferring the diverse modes (or intentions) the data displays~\citep{ghosh2023icvf}, or simply in a marginal fashion. 
    We show how each target supports different policy improvement operators, and compare both empirically.
    \item \textit{Model class.} Flow matching ~\citep{lipman2023,lipman_flow_2024} is expressive, but requires costly integration at inference. We instead investigate one-step generative modeling, focusing on drifting models ~\citep{deng2026}. Building on this framework, \textbf{we introduce a temporal difference occupancy modeling objective}, which we refer to as TD-Drift.
    \item \textit{Policy extraction.} A one-step model is cheap to differentiate, which allows extracting a policy directly from deep value gradients through the occupancy model. We compare this simple technique with the existing two-step pipeline involving critic distillation~\citep{zheng2025}.
\end{enumerate}

We systematically evaluate these design choices on four domains of the DeepMind Control Suite~\citep{tassa2018} with reward-free ExORL data~\citep{yarats2022}.
We find that (1) drifting occupancy models are competitive with flow occupancy models at a fraction of the sampling cost, (2) intention conditioning pays off, but is not strictly necessary, and (3) the direct policy extraction procedure enabled by one-step generative modeling is not only simpler, but also outperforms critic-based policy extraction.
Composing these findings, \textbf{we propose a practical method for unsupervised RL} that models the occupancy through drifting, potentially conditioned on intentions, and extracts policies directly.

In summary, our contributions are threefold: we study the essential components of the generative occupancy pipeline and offer new alternatives, we introduce a temporal-difference objective for drifting occupancy models, and we propose a simple, practical recipe based on drifting and direct policy extraction.

\section{Related Work}

\paragraph{Generative modeling in RL}
Several entities in RL, including transition kernels, policies~\citep{sutton2018}, and returns~\citep{bellemare2017distributional}, are often stochastic in nature.
In deep RL, distributions are traditionally approximated with simple families such as Gaussian policies~\citep{haarnoja2018sac, hafner2020, fujimoto2021}. Expressive generative models, on the other hand, especially diffusion and flow matching~\citep{sohl2015diffusion, ho2020ddpm, song2021sde, lipman2023, liu2023}, can now replace estimates with their diffusion-based counterparts \citep{chi2023diffusionpolicy, ding2024consistencyrl, park2025fql, dong2025}.
In model-based RL, generative transition models are unrolled for planning or imagined policy learning~\citep{heess2015, hafner2020, janner2022diffuser}, up to action-conditioned video simulators~\citep{yang2024unisim, bruce2024genie, alonso2024diamond}, but autoregressive rollouts compound errors over the horizon~\citep{talvitie2014}. Occupancy models --- the object of this work --- avoid this by predicting the discounted distribution of future states directly~\citep{janner2020}, at the cost of losing awareness of the time step at which a state is reached.
Despite the expressive power of flow models, their iterative sampling makes differentiation \emph{through} the sampler costly and unstable~\citep{park2025fql}. Since many applications, such as policy optimization via deep policy gradients, require differentiation, several workarounds have been proposed, often involving some form of distillation~\citep{park2025fql, zheng2025}.

\paragraph{One-step generative models}
Reducing the number of network evaluations of diffusion and flow models is an active line of research: via straighter probability paths~\citep{liu2023}, consistency and shortcut models~\citep{song2023, frans2025}, or mean and terminal velocities~\citep{geng2025, zhou2025}. Drifting models~\citep{deng2026} take a different route: they are one-step by construction and move the iterative transport from inference into training, by minimizing a kernel-based discrepancy between generated and data samples. We present other related works and discuss similarities in Appendix~\ref{app:extra_related}.

\paragraph{Unsupervised pre-training and occupancy models}
Reward-free offline data~\citep{yarats2022, levine2020} enables pre-training that is later adapted once rewards are revealed~\citep{ghosh2023icvf, park2024hilp, zheng2025, bagatella2025}, e.g.\ via skill discovery~\citep{eysenbach2019, park2024metra} or behavior foundation models for zero-shot RL~\citep{touati2021, touati2023, tirinzoni2025, diventura2025, urpi2025epistemically,hu2026maximum}. A central object is the successor representation~\citep{dayan1993, barreto2017} and its continuous counterpart, the occupancy or successor measure~\citep{blier2021}, on which we further comment in Appendix~\ref{app:extra_related}.

\section{Background}
\label{sec:background}

\paragraph{Markov decision processes}
We consider a reward-free Markov decision process (MDP) ${\mathcal{M} = (\mathcal{S}, \mathcal{A}, p, \gamma, \mu_0)}$ with state space $\mathcal{S}$, action space $\mathcal{A}$, transition kernel $p(s' \mid s, a)$, 
discount factor $\gamma \in [0, 1)$ and initial state distribution $\mu_0$. A policy $\pi(\cdot \mid s)$ is a conditional distribution over actions given a state.
We consider a task to be specified by a reward function $r: \mathcal{S} \to \mathbb{R}$, turning $\mathcal{M}$ into a standard MDP $\mathcal{M}_r = (\mathcal{S}, \mathcal{A}, p, \gamma, r, \mu_0)$.
The goal of RL is to find a policy that maximizes the expected discounted cumulative reward $J(\pi) = \mathbb{E}_{\pi}\big[\sum_{t=0}^\infty \gamma^t r(s_{t+1})\big]$, where $s_0 \sim \mu_0$, $s_{t+1} \sim p(\cdot \mid s_t, a_t)$ and $a_t \sim \pi(\cdot \mid s_t)$.

\paragraph{Occupancy models}
The occupancy (or successor measure) of a policy $\pi$ is the discounted distribution over future states when starting in $s$, taking action $a$, and following $\pi$ thereafter~\citep{dayan1993, blier2021}:
\begin{equation}
    m^\pi(S^+ \mid s, a) = (1 - \gamma) \sum_{t=0}^\infty \gamma^t\, \mathrm{Pr}(S_{t+1} \in S^+ \mid S_0 = s, A_0 = a, \pi),
    \label{eq:occupancy}
\end{equation}
with $S^+ \subseteq \mathcal{S}$ and $\mathrm{Pr}(S_{t+1} \in S^+ \mid S_0, A_0, \pi)$ as the probability of reaching a state in $S^+$ at step $t+1$. The occupancy satisfies a Bellman equation~\citep{janner2020}:
\begin{equation}
    m^\pi(\cdot \mid s, a) = (1-\gamma)\, p(\cdot \mid s, a) + \gamma\, \mathbb{E}_{s' \sim p(\cdot \mid s, a),\, a' \sim \pi(\cdot \mid s')}\big[m^\pi(\cdot \mid s', a')\big].
    \label{eq:bellman_occupancy}
\end{equation}
Finally, the expected discounted sum of rewards collected from a given state and action and following policy $\pi$ thereafter (i.e., the action-value function) can be expressed as
\begin{equation}
    Q^\pi_r(s, a) = \frac{1}{1-\gamma}\, \mathbb{E}_{s^+ \sim m^\pi(\cdot \mid s, a)}\big[r(s^+)\big],
    \label{eq:value_decomposition}
\end{equation}
which illustrates how the occupancy model enables evaluating $\pi$ for arbitrary tasks.

\paragraph{Flow matching and drifting}
Flow matching~\citep{lipman2023, liu2023} learns a velocity field $v_\theta(x_t, t)$ that transports noise $x_0 \sim \mathcal{N}(0, \text{I})$ to data $x_1$ by regressing onto $x_1 - x_0$ at interpolants $x_t = (1-t)\,x_0 + t\,x_1$. Samples are obtained by integrating the ODE $\dot x_t = v_\theta(x_t, t)$ from $t=0$ to $t=1$, which requires many network evaluations. Drifting models~\citep{deng2026}, instead, train a one-step generator $f_\theta: \epsilon \mapsto x$, $\epsilon \sim \mathcal{N}(0, \text{I})$, whose pushforward distribution $q_\theta$ should match a target distribution $p$. Generated samples $x$ are moved by a \emph{drifting field} $V_{p,q}$ that attracts them towards positive samples $y^+ \sim p$ and repels them from negative samples $y^- \sim q_\theta$,
\begin{equation}
    V_{p,q}(x) = \underbrace{\frac{\mathbb{E}_{y^+ \sim p}\big[k(x, y^+)\,(y^+ - x)\big]}{\mathbb{E}_{y^+ \sim p}\big[k(x, y^+)\big]}}_{V^+_p(x)} - \underbrace{\frac{\mathbb{E}_{y^- \sim q_\theta}\big[k(x, y^-)\,(y^- - x)\big]}{\mathbb{E}_{y^- \sim q_\theta}\big[k(x, y^-)\big]}}_{V^-_q(x)},
    \label{eq:drift_field}
\end{equation}
with a kernel $k$, e.g., the exponential kernel $k(x, y) = \exp(-\|x - y\|/\kappa)$ with temperature $\kappa$. The generator $f_\theta$ is trained so its output is a self-consistent fixed point of the field $V_{p, q}$, via
\begin{equation}
    \mathcal{L}_{\mathrm{Drift}}(\theta) = \mathbb{E}_{\epsilon \sim \mathcal{N}(0, \text{I})}\Big[\big\|f_\theta(\epsilon) - \mathrm{sg}\big(f_\theta(\epsilon) + V_{p, q}(f_\theta(\epsilon))\big)\big\|^2\Big],
    \label{eq:drift_loss}
\end{equation}
where $\mathrm{sg}$ denotes the stop-gradient operator, such that $q_\theta = p \Rightarrow \|V_{p, q}\| = 0$. 

\section{Methods}
\label{sec:methods}

We study a pipeline for unsupervised RL which operates in two stages following~\citet{zheng2025}.
During \emph{pre-training}, a large reward-free dataset of transitions $\mathcal{D} = \{(s, a, s', a')\}$ is available; we will fit several variants of occupancy models on $\mathcal{D}$ (see \Cref{sec:methods_model}).
During \emph{policy extraction}, a reward-labeled dataset $\mathcal{D}_r = \{(s, a, r)\}$ from a downstream task is exposed, and we learn a reward-maximizing policy $\pi_\omega$ on top of the frozen occupancy model. 
In this phase, we fit a reward model $r_\phi: \mathcal{S} \rightarrow \mathbb{R}$ on $\mathcal{D}_r$,
and study different techniques to extract a policy (see \Cref{sec:methods_extraction}).

We will focus on three design choices at the core of this pipeline: (i) \emph{what to fit} -- whether the different modes composing an occupancy are explicitly inferred and modeled (\Cref{sec:methods_conditional}), (ii) \emph{how to fit it} -- the class of generative model used to represent the occupancy (\Cref{sec:methods_model}), and (iii) \emph{how to extract a policy} from the occupancy once a reward is provided (\Cref{sec:methods_extraction}). 
For each of these decisions, we review existing solutions and present novel alternatives or simplifications.

\subsection{Conditional or Unconditional Modeling}
\label{sec:methods_conditional}
We model occupancies by learning a generative sampler $s^+ \sim \hat m^\pi(\cdot \mid s, a) \approx m^\pi(\cdot \mid s,a)$. Here, the superscript $\pi$ denotes the data-generating policy whose occupancy is modeled, i.e., the behavior policy $\pi_\beta$ or an intention-conditioned policy $\pi_z$, and not the extracted policy $\pi_\omega$.
Occupancy models are often fitted on data collected by a mixture of policies. 
This raises a design choice: the degree to which the multimodal structure of the data must be captured by the occupancy model. 
Prior work has addressed this by learning occupancy models for a family of policies $\pi_z$ indexed by a latent $z$ ~\citep{barreto2017, touati2021}. InFOM ~\citep{zheng2025} infers intentions ${z \sim p_e(\cdot \mid s', a')}$ from consecutive transitions ${(s,a,s',a')}$ via a variational encoder, and predicts the occupancy measure of future states, $m^{\pi_z}(\cdot \mid s,a)$, based on that intention. 
Importantly, the occupancy model and the intention encoder are optimized jointly via the following objective:
\begin{equation}
    \mathcal{L}_{\mathrm{cond}}(\theta, e) = \mathcal{L}^{\pi_z}_{\mathrm{Flow}\mid\mathrm{Drift}}(\theta) + \lambda_{\mathrm{KL}} \mathbb{E}_{(s',a') \sim \mathcal{D}}\big[D_{\mathrm{KL}}\big(p_e(\cdot \mid s', a') \,\|\, \mathcal{N}(0, \text{I})\big)\big].
    \label{eq:conditional_loss}
\end{equation}
Here, $z\sim p_e(\cdot \mid s', a')$, and $\mathcal{L}^{\pi_z}_{\mathrm{Flow}\mid\mathrm{Drift}}$ denotes the occupancy model's training loss, which takes a different form depending on the underlying generative model class, as described in \Cref{sec:methods_model}.

\paragraph{Unconditional modeling} We explore a simpler, \emph{unconditional} alternative, which drops the encoder together with the KL term, and only minimizes $\mathcal{L}^{\pi_\beta}_{\mathrm{Flow}\mid\mathrm{Drift}}(\theta)$ such that it fits the occupancy $m^{\pi_\beta}$ of the marginal behavior policy $\pi_\beta$ that generated the dataset $\mathcal{D}$.
Despite a conditional model being more expressive (and enabling a different form of policy extraction, as we will see in Section~\ref{sec:methods_extraction}), jointly learning the occupancy model and a policy latent as in \Cref{eq:conditional_loss} can be very challenging, since it adds complexity to the pipeline, and its benefits might be offset by poor intention inference.
\subsection{Model Class}
\label{sec:methods_model}

Occupancies in continuous spaces can be very complex, which calls for an expressive model class, such as flow-based generative models. 
TD-Flow~\citep{farebrother2025, zheng2025} proposes a training objective that enforces the Bellman equation (\Cref{eq:bellman_occupancy}) at the level of probability paths, mixing a next-state (\Cref{eq:td_flow_next}) and a bootstrap (\Cref{eq:td_flow_bootstrap}) regression term. 
Given a transition $(s, a, s', a') \sim \mathcal{D}$, noise $\epsilon \sim \mathcal{N}(0, \text{I})$, and time $t \sim \mathcal{U}[0, 1]$, the next-state term regresses onto the straight path towards $s'$, while the bootstrap term regresses onto the target velocity field  $v^\pi_{\bar \theta}$ at a future state $\bar s^+ \sim \bar m^\pi(\cdot \mid s', a')$, obtained by integrating $v^\pi_{\bar \theta}$ at the next state-action pair $(s', a')$.
\begin{align}
    \mathcal{L}^\pi_{\mathrm{Flow}}(\theta) &= (1 - \gamma)\, \mathcal{L}^\pi_{\mathrm{next}}(\theta) + \gamma\, \mathcal{L}^\pi_{\mathrm{boot}}(\theta),
    \label{eq:td_flow}
    \\ \mathcal{L}^\pi_{\mathrm{next}}(\theta) &= \mathbb{E}\Big[\big\| v_\theta^\pi\big(x_t, t \mid s, a\big) - (s' - \epsilon) \big\|^2\Big], \quad & x_t &= (1-t)\epsilon + t s' \label{eq:td_flow_next}
    \\ \mathcal{L}^\pi_{\mathrm{boot}}(\theta) &= \mathbb{E}\Big[\big\| v_\theta^\pi\big(\bar x_t, t \mid s, a\big) - \mathrm{sg}\big(v^\pi_{\bar \theta}(\bar x_t, t \mid s', a')\big) \big\|^2\Big], \quad & \bar x_t &= (1-t)\epsilon + t \bar s^+ \label{eq:td_flow_bootstrap}
\end{align}
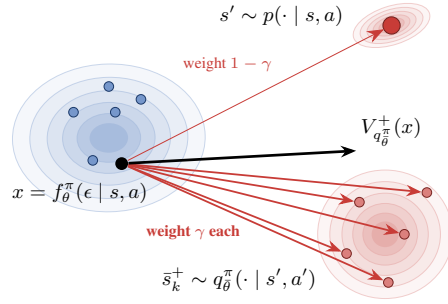
\begin{wrapfigure}[18]{r}{2.75in}
    \centering
    \scalebox{0.85}{
\newcommand{\dhill}[6]{%
  \begin{scope}[shift={(#2,#3)}, rotate=#6]
    \foreach \s/\op in {1.0/0.05, 0.82/0.06, 0.64/0.07, 0.48/0.09, 0.33/0.11, 0.2/0.13}{
      \fill[#1, opacity=\op] (0,0) ellipse ({\s*#4} and {\s*#5});
    }
    \foreach \s in {1.0,0.82,0.64,0.48,0.33}{
      \draw[#1!45, line width=0.2pt, opacity=0.55] (0,0) ellipse ({\s*#4} and {\s*#5});
    }
  \end{scope}%
}

\begin{tikzpicture}[
    >={Stealth[length=2.4mm]},
    font=\small,
    boot/.style={fill=rot!65, draw=rot!60!black, circle, inner sep=1.5pt}, 
    nxt/.style={fill=rot, draw=rot!55!black, circle, inner sep=2.7pt},      
    aw/.style={->, rot, line width=#1},                                     
]
\dhill{blau}{-0.0}{0}{1.5}{1.15}{1.5}          
\dhill{rot}{4.4}{1.75}{0.62}{0.3}{18}     
\dhill{rot}{4.3}{-1.5}{1.05}{1.0}{0}      

\node[boot, fill=blau!65, draw=blau!60!black] (g1) at (-0.55,0.4) {};
\node[boot, fill=blau!65, draw=blau!60!black] (g2) at (0.1,0.4) {};
\node[boot, fill=blau!65, draw=blau!60!black] (g3) at (-0.25,-0.35) {};
\node[boot, fill=blau!65, draw=blau!60!black] (g4) at (0.0,0.8) {};
\node[boot, fill=blau!65, draw=blau!60!black] (g5) at (0.5,0.6) {};

\node[circle, fill=black, inner sep=2pt,
   label={[font=\footnotesize]west:}] (x) at (0.2,-0.4) {};
\node[font=\footnotesize] at (-0.45,-0.85) {$x=f_\theta^\pi(\epsilon \mid s, a)$};

\node[nxt] (sp) at (4.4,1.75) {};
\node[anchor=west, font=\footnotesize] at (1.6,1.9)  {$s'\sim p(\cdot\mid s,a)$};
\draw[aw=0.2pt] (x) -- (sp);
\node[font=\scriptsize, text=rot, fill=white, fill opacity=0.0, text opacity=1, inner sep=1.5pt] at (1.85,1.15) {weight $1-\gamma$};

\node[boot] (b1) at (3.9,-1.0) {};
\node[boot] (b2) at (4.6,-1.5) {};
\node[boot] (b3) at (3.7,-1.8) {};
\node[boot] (b4) at (4.95,-0.85) {};
\node[boot] (b5) at (4.35,-2.25) {};
\node[anchor=west, font=\footnotesize] at (0.7,-2.2) {$\bar s_k^+\sim q_{\bar\theta}^\pi(\cdot\mid s', a')$};
\foreach \b in {b1,b2,b3,b4,b5}{ \draw[aw=0.8pt] (x) -- (\b); }
\node[font=\scriptsize, text=rot, fill=white, fill opacity=0.0, text opacity=1, inner sep=1.5pt, anchor=north east] at (2.1,-1.3) {\textbf{weight} $\mathbf{\gamma}$ \textbf{each}};

\draw[->, very thick, black] (x) -- (3.85,-0.2) node[anchor=south west, font=\footnotesize\bfseries, inner sep=1.5pt] {$V^+_{q_{\bar\theta}^\pi}(x)$};
\end{tikzpicture}}
    \vspace{-0mm}
    \caption{\textbf{Attracting TD-Drift field} $\mathbf{V^+_{q_{\bar\theta}^\pi}}$. Each sample from the pushforward distribution (blue) is attracted with weight $1 - \gamma$ (light red) towards the next state $s'$, while the attraction towards the bootstrap samples $\bar s^+$ (bold red) is weighted with $\gamma$.\looseness -1}
    \vspace{-5mm}
 \label{fig:positive_drift_field}
\end{wrapfigure}
Flow matching has proven very powerful and precise for generating individual samples, but sample quality depends on the number of integration steps, and thus on inference compute~\citep{frans2025}. 
In occupancy learning, this cost is paid twice: computing every bootstrap target $\bar s^+$ (used for training the occupancy model) and every Monte Carlo value estimate (used for policy extraction, see \Cref{sec:methods_extraction}), both of which require a full ODE integration.
Motivated by this limitation, we introduce a Bellman-consistent one-step sampler, which we call \emph{TD-Drift}.\looseness -1

\paragraph{TD-Drift}
We build on drifting (Equations \ref{eq:drift_field} and \ref{eq:drift_loss}) with a conditional generator $f_\theta^\pi(\epsilon \mid s, a)$ whose pushforward distribution $q_\theta^\pi(\cdot \mid s, a)$ approximates $m^\pi(\cdot \mid s, a)$. Drifting requires samples from the target distribution -- in this case, the measure $m^\pi$. 
Based on the ideas from \citet{farebrother2025,zheng2025}, on the Bellman equation (\ref{eq:bellman_occupancy}), and on temporal difference learning, we instantiate a training objective for drifting occupancy models. We write $m^\pi$ at $(s,a)$ as a mixture between the next state distribution $p$, and the measure at the next state, in practice estimated through $q^\pi_{\bar\theta}$, a target copy of the model:
\begin{equation}
    m^\pi(\cdot \mid s, a) \approx (1-\gamma)\, p(\cdot \mid s, a) + \gamma\, \mathbb{E}_{s' \sim p(\cdot \mid s, a),\, a' \sim \pi(\cdot \mid s')}\big[q^\pi_{\bar\theta}(\cdot \mid s', a')\big].
    \label{eq:td_drift_target}
\end{equation}
By linearity of expectation, the attraction field in \Cref{eq:drift_field} towards this mixture becomes:
\begin{equation}
    V^+_{q_{\bar\theta}^\pi}(x) = \frac{(1 - \gamma)\, \mathbb{E}_{s'}\big[k(x, s')(s' - x)\big] + \gamma\, \mathbb{E}_{\bar s^+, s', a'}\big[k(x, \bar s^+)(\bar s^+ - x)\big]}{(1 - \gamma)\, \mathbb{E}_{s'}\big[k(x, s')\big] + \gamma\, \mathbb{E}_{\bar s^+, s', a'}\big[k(x, \bar s^+)\big]},
    \label{eq:td_attraction}
\end{equation}
with $s' \sim p(\cdot \mid s, a)$ and $\bar s^+ \sim q_{\bar\theta}^\pi(\cdot \mid s', a')$, $a' \sim \pi(\cdot \mid s')$. See \Cref{app:impl_details} for more details.

Inserting the resulting field into \Cref{eq:drift_loss} yields the TD-Drift loss for policy $\pi$,
\begin{equation}
    \mathcal{L}^\pi_{\mathrm{Drift}}(\theta) = \mathbb{E}_{(s,a,s',a') \sim \mathcal{D}, x \sim q_\theta^\pi(\cdot \mid s, a)}\Big[\big\|x - \mathrm{sg}\big(x + V^+_{q_{\bar\theta}^\pi}(x) - V^-_q(x)\big)\big\|^2\Big],
    \label{eq:td_drift}
\end{equation}
which plays the same role as $\mathcal{L}^\pi_{\mathrm{Flow}}$ in \Cref{eq:td_flow}.
We visually describe the attractive component of this objective in \Cref{fig:positive_drift_field}.
In contrast to TD-Flow, TD-Drift generates bootstrap samples from the estimated measure at the next state, required in~\Cref{eq:td_attraction}, with a single forward pass of the target network.

\subsection{Policy Extraction}

\label{sec:methods_extraction}
The third design choice we explore concerns policy extraction, once a reward-labeled dataset is available. The implementation of this step depends predominantly on the multimodality of the modeled occupancy (\Cref{sec:methods_conditional}) and on the generative model class used to fit it (\Cref{sec:methods_model}). We first address the case of more general conditional occupancy models, and later describe a lighter unconditional variant.

A conditional model estimates a family of occupancies, allowing for policy extraction via generalized policy improvement (GPI) ~\citep{barreto2017}. 
Intuitively, GPI optimizes not only over actions but also over intentions, i.e., $\argmax_\omega \mathbb{E}_{a\sim\pi_\omega(\cdot \mid s)}\max_{z} Q_{z}(s,a)$, where each intention-conditioned value (\Cref{eq:value_decomposition}) can be estimated by Monte Carlo (MC) via:
\begin{equation}
    \hat Q_z(s,a)=  \frac{1}{(1-\gamma)N}\sum_{i=1}^N r_\phi(s^+_i), \quad s^+_i \sim \hat m^{\pi_z}(\cdot \mid s,a).
\label{eq:q_mc}
\end{equation}
InFOM first distills these estimates into an explicit critic $Q_\psi$ via:
\begin{equation}
    \mathcal{L}_Q(\psi) = \mathbb{E}_{(s,a) \sim \mathcal{D}_r,\, z \sim p(z)}\Big[\ell^\tau_2\Big(\hat Q_z(s,a) - Q_\psi(s, a)\Big)\Big],
    \label{eq:expectile_critic}
\end{equation}
where $p(z)=\mathcal{N}(0,\text{I})$, and the expectile loss $\ell^\tau_2(u) = |\tau - \mathbf{1}(u < 0)|\, u^2$~\citep{kostrikov2021} with $\tau \in (0.5, 1)$ regresses $Q_\psi$ towards an upper expectile over intentions, implicitly approximating the maximum over $z$. 
The policy $\pi_\omega$ is then extracted with a behavior-regularized DDPG-BC objective~\citep{fujimoto2021}, $\mathcal{L}_\pi(\omega) = \mathbb{E}_{(s, a) \sim \mathcal{D}_r,\, \hat a \sim \pi_\omega(\cdot \mid s)}\big[-Q_\psi(s, \hat a) - \alpha \log \pi_\omega(a \mid s)\big]$ with BC weight $\alpha$, following the action gradient $\nabla_a Q_\psi(s,a)$ through the distilled critic (see Appendix~\ref{app:impl_details} for more details).
Crucially, this two-step procedure avoids backpropagating through the occupancy model, which is expensive and reported to be unstable for flow-based models ~\citep{zheng2025}.

In contrast to flow-based models, drifting models generate samples with a single forward pass. We hypothesize that their gradients with respect to the action are informative enough to extract a policy directly, and hence propose a novel policy extraction method that circumvents the need to train an additional critic network.

\paragraph{Direct extraction}
With drifting models, future states in \Cref{eq:q_mc} are sampled as ${s^+_i = f_\theta(\epsilon_i \mid s, a, z)}$ with $\epsilon_i \sim \mathcal{N}(0, \text{I})$, so the MC estimate is itself differentiable with respect to the action. Instead of approximating the maximum over intentions implicitly, we perform a zero-order search over a finite set of $M$ candidate intentions, i.e., ${z^*(s,a) = \argmax_{z \in \{z_1, \ldots, z_M\}} \hat Q_z(s, a)}$, where $z_m \sim \mathcal{N}(0, \text{I})$. Treating $z^*$ as fixed,
the action gradient $\nabla_a \hat Q$ follows from the chain rule, composing the gradient of the reward model $r_\phi$ with the Jacobian of the generator $f_\theta$:
\begin{equation}
    \nabla_a \hat Q_{z^*}(s, a) = \frac{1}{1 -\gamma}\,\frac{1}{N} \sum_{i = 1}^N \frac{\partial r_\phi(s^+_i)}{\partial s^+_i}\, \frac{\partial f_\theta(\epsilon_i \mid s, a,z^*)}{\partial a}.
    \label{eq:mc_gradient}
\end{equation}
This gradient replaces $\nabla_aQ_\psi$ in the DDPG-BC objective of $\pi_\omega$, removing the need for a distilled critic. Optimizing the intention with first-order methods would also be a viable approach, but we do not consider it in this work.
\paragraph{A note on unconditional extraction}
While we have described extraction for conditional occupancy models, unconditional variants yield an even lighter pipeline, since no optimization over intentions is needed. In this case, GPI reduces to a single step of policy improvement of the behavior value function~\citep{brandfonbrener2021onestep}.
For critic distillation, the expectile loss in \Cref{eq:expectile_critic} simplifies to standard MSE ($\tau=0.5$), while for direct policy extraction, no search is needed and \Cref{eq:mc_gradient} is used as is, with the dependence on $z$ dropped.

\section{Experiments}
\label{sec:experiments}

\paragraph{Methods}
Crossing the three decisions of \Cref{sec:methods} yields six variants of occupancy modeling (OM) depending (i) on whether they are conditioned on \textit{In}tentions or not, (ii) on the model (\textit{F}low or \textit{D}rift), and (iii) on whether policy extraction is \textit{Direct}: InFOM, FOM, InDOM, DOM, InDOM-Direct and DOM-Direct.
We do not consider direct extraction for flow models, since it would require differentiating through the ODE solver.
The first and the second combinations (InFOM and FOM) recover methods introduced by ~\citet{zheng2025}, while the remaining ones are novel instantiations. 
All share the same reward and policy model, so they differ only in the studied components; the hyperparameter tuning protocol is reported in Appendix~\ref{app:hyperparameters}. 
As references, we also report behavior cloning (BC) as a lower bound, and forward-backward representations (FB)~\citep{touati2023} rerun from \citet{bagatella2025}, which operate in different frameworks and are therefore shown in gray.

\paragraph{Domains, data and protocol}
We evaluate on four domains of the DeepMind Control Suite~\citep{tassa2018}: \texttt{cheetah}, \texttt{walker}, \texttt{quadruped}, and \texttt{jaco}, each spanning four tasks that share dynamics and differ only in the reward.
We will largely report performance averaged over the tasks, and refer to~\Cref{tab:task_results} for detailed results.
Following \citet{zheng2025}, occupancy models are pre-trained on a reward-free exploratory dataset with $1$M transitions from ExORL~\citep{yarats2022}, gathered by random network distillation \citep{burda2019rnd}.
The occupancy model is then frozen, and the policy is extracted on a reward-labeled dataset with $500$K transitions of the same domain.
We report the undiscounted episode return of the extracted policy averaged over $50$ evaluation episodes and four tasks, as mean $\pm$ standard error (s.e.m.) over five independent seeds unless stated otherwise (see Appendix~\ref{app:evaluation_protocol}).

We will first seek a direct empirical answer to the three design choices we highlighted in ~\Cref{sec:methods}, and then provide a broader comparison between the best recipe and existing methods, and ablate its most important components.

\begin{wrapfigure}[14]{r}{2.75in}
    \centering
    \vspace{-4mm}
    \includegraphics[width=\linewidth]{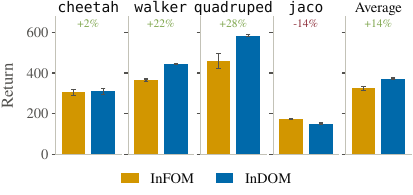}
    \vspace{-7mm}
    \caption{\textbf{Drifting vs.\ flow matching.} Return of InFOM and InDOM per domain (mean $\pm$ s.e.m.). The percentage is the performance difference of InDOM relative to InFOM.\looseness-1}
    \vspace{-4mm}
    \label{fig:Q1InFOMvsInDOM}
\end{wrapfigure}

\subsection{Is drifting an effective solution to occupancy modeling?}
\label{sec:q1}

\Cref{fig:Q1InFOMvsInDOM} compares an existing, flow-based pipeline (InFOM) with a variant that replaces the occupancy model with one based on drifting (InDOM), adopting the training objective we derived in Section~\ref{sec:methods_model}.
Both are conditional and extract a policy from a distilled critic, so that they differ only in the generative model.
Overall, InDOM outperforms InFOM, particularly on \texttt{quadruped}, where the flow model is also markedly less stable across seeds. InFOM remains ahead in \texttt{jaco}.
Averaged over the four domains, InDOM reaches an improvement of $14\%$: drifting is thus a competitive replacement for flow matching as an occupancy model, while requiring a single network evaluation per generation instead of an ODE integration (see~\Cref{fig:runtime_bars_mean} for a wall-clock comparison).

\begin{wrapfigure}{r}{1.8in}    
    \centering
    \vspace{-8mm}
    \includegraphics[width=\linewidth]{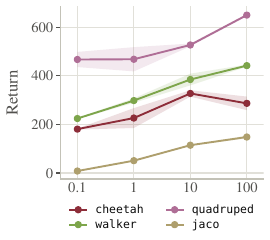}
    \vspace{-7mm}
    \caption{\textbf{KL weight of the intention encoder.} Return of InDOM over the KL weight. Each sweep uses a fixed configuration per domain. (mean $\pm$ s.e.m. over three random seeds).\looseness-1}
    \vspace{-4mm}
    \label{fig:Q2_KL_ablations}
\end{wrapfigure}

\subsection{Is conditional occupancy modeling necessary?}
\label{sec:q2}

The degree to which different intentions can be modeled is governed by the weight $\lambda_{\mathrm{KL}}$ of the KL term that regularizes the intention posterior towards its prior (see Equation~\ref{eq:conditional_loss}).
Sweeping its values in Figure~\ref{fig:Q2_KL_ablations}, we observe that in every domain the return of InDOM grows with $\lambda_{\mathrm{KL}}$.
Modeling a conditional occupancy is therefore not free: an under-regularized intention encoder, which in principle should extract more expressive intentions, in practice actively harms the downstream policy.

\begin{figure}[t]
    \centering
    \includegraphics[width=\linewidth]{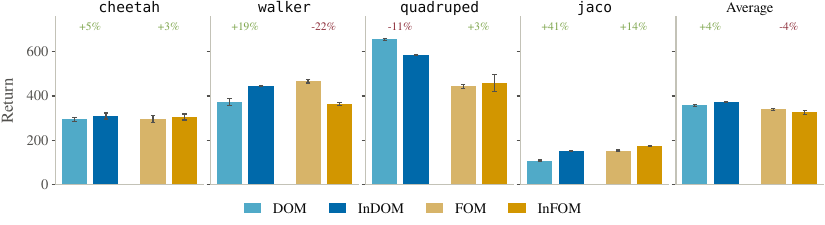}
    \vspace{-7mm}
    \caption{\textbf{Effect of intention conditioning.} Return with and without intention conditioning for the drift (DOM, InDOM) and the flow (FOM, InFOM) occupancy models. The percentage is the change in performance due to conditional measure modeling. (mean $\pm$ s.e.m.)}
    \label{fig:Q2_intention_ablation}
\end{figure}

A latent that is pushed this close to its prior invites the hypothesis that it carries almost no information at all, such that an unconditional model should do at least as well.
\Cref{fig:Q2_intention_ablation} hints that this may often be the case: at its best KL weight, conditioning on intentions can still raise performance for both InFOM and InDOM, but unconditional variants (FOM and DOM) are overall competitive.
While intention modeling can be beneficial, performance gains are modest, require a carefully tuned KL regularization, and the instantiation of an additional encoder, as discussed in Section~\ref{sec:methods_conditional}.\looseness -1

\subsection{Is direct policy extraction competitive?}
\label{sec:q3}

\begin{figure}
    \centering
    \includegraphics[width=\linewidth]{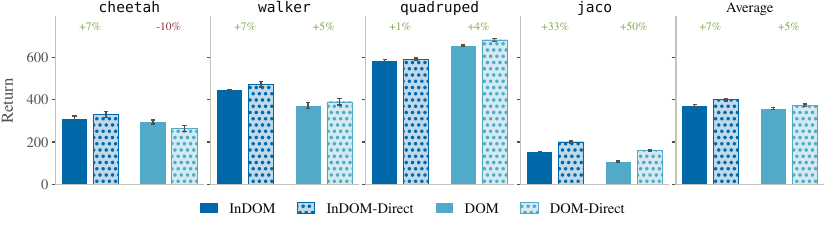}
    \vspace{-7mm}
    \caption{\textbf{Critic-based vs.\ direct extraction.} Return of InDOM and DOM with policy extraction through a distilled critic and direct policy extraction. The percentage is the 
    performance change of direct policy extraction over extraction from a distilled critic. (mean $\pm$ s.e.m.)}
    \label{fig:Q3_policy_extraction}
\end{figure}

Since a drift model generates future states in a single forward pass, gradients from the Monte Carlo value estimates should be informative and therefore make direct policy extraction, without training a critic (Equation~\ref{eq:mc_gradient}), possible.
\Cref{fig:Q3_policy_extraction} compares both routes for InDOM and DOM.
We find that direct extraction not only simplifies the occupancy modeling pipeline (removing the need for critic distillation), but it also matches or improves over critic-based extraction in seven of eight settings, by up to $50\%$.

\begin{wrapfigure}{r}{1.8in}
    \centering
    \vspace{-4mm}
    \includegraphics[width=\linewidth]{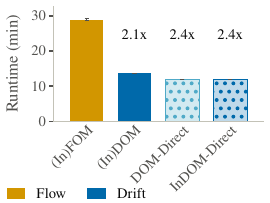}
    \vspace{-7mm}
    \caption{\textbf{Wall-clock time for policy extraction}. Extrapolated: $10$K to $500$K steps. (mean $\pm$ s.e.m.).}
    \vspace{-4mm}
    \label{fig:runtime_bars_mean}
\end{wrapfigure}

\subsection{Is Policy Extraction via drift occupancies more efficient?}
\label{sec:q3b}

Drifting is by design a single neural-function-evaluation (NFE) sampler. 
Since we compare it throughout this work with a flow-based sampler, which uses $10$ integration steps, we also benchmark both approaches with respect to their inference cost (over $500$K steps) and report results in \Cref{fig:runtime_bars_mean}. 
Flow-based methods are expensive, especially due to critic distillation, which requires sampling through the ODE solver for every target critic sample (see~\Cref{eq:q_mc}). InFOM and FOM methods incur the same cost, since they extract the policy through a similar distilled critic, and intention conditioning does not add overhead during this process.  
Drifting models are generally cheaper: for InDOM and DOM, critic distillation is cheaper, as it only requires a single NFE to compute targets.
Finally, InDOM-Direct and DOM-Direct are even faster, as they do not need to distill a critic at all, yielding a speedup of  $2.4\times$ over (In)FOM. 
On top of this, we find that the cost of zero-order optimization over intentions required by InDOM-Direct is almost negligible.

\subsection{What matters for occupancy modeling and downstream policy extraction?}
\label{sec:q4}

Previous evaluations hinted that the choice of generative model and extraction method matters; modeling intentions does too, albeit to a lesser extent.
We thus highlight two recipes and compare them with existing methods in \Cref{tab:main}.
Both InDOM-Direct and DOM-Direct simplify the occupancy modeling pipeline by fitting a drifting model through which the policy is extracted directly.
They thus remove the need for a critic; DOM-Direct also needs no intention encoder, no KL weight, and no optimization over latent candidates.
DOM-Direct outperforms InFOM on \texttt{quadruped} and \texttt{walker}, and stays close on \texttt{cheetah} and \texttt{jaco}, on average scoring $375$ against $326$.
Adding intentions back, InDOM-Direct is the strongest variant overall.
All occupancy-based methods clearly exceed behavior cloning, while FB, which does not rely on generative modeling and adopts a fundamentally different approach, remains stronger on \texttt{cheetah} and \texttt{walker} but falls behind on \texttt{quadruped} and \texttt{jaco}.

\begin{table}[t]
    \centering
    \caption{\textbf{Comparison with existing methods.} Four-task mean return per domain (mean $\pm$ s.e.m.\ over five pre-training seeds), each method at its best discount and configuration per domain. \textcolor{gray}{FB} are rerun using the codebase of \citet{bagatella2025}. Per-task numbers and the remaining variants are in \Cref{tab:task_results}.}
    \label{tab:main}
    \vspace{4mm}
\begin{tabular}{l r@{\,{\scriptsize$\pm$}\,}l r@{\,{\scriptsize$\pm$}\,}l r@{\,{\scriptsize$\pm$}\,}l r@{\,{\scriptsize$\pm$}\,}l r@{\,{\scriptsize$\pm$}\,}l r@{\,{\scriptsize$\pm$}\,}l}
\toprule
Environment & \multicolumn{2}{c}{BC} & \multicolumn{2}{c}{\textcolor{gray}{FB}} & \multicolumn{2}{c}{InFOM} & \multicolumn{2}{c}{DOM-Direct} & \multicolumn{2}{c}{InDOM-Direct} \\
\midrule
\texttt{cheetah} & $142$ & {\tiny$2$} & \textcolor{gray}{$631$} & \textcolor{gray}{{\tiny$16$}} & $305$ & {\tiny$13$} & $266$ & {\tiny$14$} & $\mathbf{331}$ & {\tiny$14$}  \\
\texttt{walker} & $232$ & {\tiny$2$} & \textcolor{gray}{$780$} & \textcolor{gray}{{\tiny$12$}} & $364$ & {\tiny$7$}  & $390$ & {\tiny$15$} & $\mathbf{474}$ & {\tiny$13$} \\
\texttt{quadruped} & $501$ & {\tiny$3$} & \textcolor{gray}{$601$} & \textcolor{gray}{{\tiny$18$}} & $458$ & {\tiny$37$} & $\mathbf{682}$ & {\tiny$6$} & $592$ & {\tiny$6$} \\
\texttt{jaco} & $3$ & {\tiny$1$} & \textcolor{gray}{$2$} & \textcolor{gray}{{\tiny$1$}} & $175$ & {\tiny$3$} & $161$ & {\tiny$3$} & $\mathbf{201}$ & {\tiny$5$} \\
\midrule
\textbf{Mean} & \multicolumn{2}{l}{$220$} & \multicolumn{2}{l}{\textcolor{gray}{$503$}} & \multicolumn{2}{l}{$326$} & \multicolumn{2}{l}{$375$} & \multicolumn{2}{l}{$\mathbf{400}$} \\
\bottomrule
\end{tabular}

\end{table}

\section{Discussion}
\label{sec:discussion}
The combination of occupancy modeling with downstream policy extraction constitutes a promising technique for unsupervised RL.
We decompose this recipe along three axes: what is modeled (conditional or marginal occupancies), how it is modeled (iterative or one-step generative models), and how a policy is extracted (via a distilled critic or directly).
We propose simpler alternatives to prior choices on each axis and introduce TD-Drift, a temporal difference objective for drifting models of occupancies, which allows direct policy extraction. When tested across several domains, methods based on TD-Drift outperform the existing flow-based pipeline while being conceptually simpler and computationally more efficient.

\textbf{Limitations\;} Our move towards drifting trades cheaper inference for more training compute: distributional distances based on samples (drift) are generally more expensive than matching vector fields (flow). Our method thus requires a longer pre-training phase for learning the occupancy model. However, this cost is paid offline and amortized over all downstream tasks. In exchange, our method is more efficient when it matters most, at inference time.

As we followed the standard benchmark procedure for unsupervised reinforcement learning, we focused on datasets collected through self-supervised exploration, and we did not evaluate other sources of data, such as expert demonstrations. Finally, we measured \emph{whether} intentions help but did not inspect \emph{what} the intention manifold encodes, which leaves the question of the reason for the performance gap open.

\textbf{Future work\;}
Our pipeline evaluated a two-step policy extraction technique based on critic distillation, and a simple, direct policy extraction objective. These are, in practice, two ends of a spectrum of methods for policy optimization from a learned occupancy: first-order and zero-order approaches can be instantiated and combined to further improve policy learning.
More generally, investigating the scalability of drifting-based occupancy modeling by defining kernels in latent space is an interesting direction, as demonstrated in image generation \citep{deng2026}. Learning the latent space in which this should occur is even more exciting. 

\newpage

\paragraph{Acknoledgements}
We would like to thank the members of the Autonomous Learning Group for the many thoughtful discussions that shaped this work
Núria Armengol Urpi and Marco Bagatella are supported by the Max Planck ETH Center for Learning Systems.
Georg Martius is a member of the Machine Learning Cluster of Excellence, EXC number 2064/1 -- Project number 390727645.

\paragraph{Reproducibility statement}
To ensure reproducibility of our results, we make our code publicly available on our \href{https://sites.google.com/view/td-drift/}{website}. We additionally provide all relevant hyperparameters, details for training and evaluating each method, and access to external data sources in~\Cref{sec:experiments} and Appendix~\ref{app:hyperparameters} to ~\ref{app:impl_details}.

\paragraph{Ethics statement}
Our work adheres to the ICLR Code of Ethics. It does not involve human subjects, and it raises no privacy, security, or other ethical concerns that we are aware of. Our work does focus on reinforcement learning for embodied control, and we recognize that autonomous robotic systems could be deployed in harmful applications. However, our contributions are fundamental and algorithmic in nature, and they do not directly target any specific negative use case.

\paragraph{AI disclosure}

In this work, we used generative AI tools for refining hypotheses, providing feedback on research methodology and experiments, and coding assistance for implementing the methods. 
We did not use generative AI tools for generating synthetic datasets, developing theoretical models or conceptual frameworks, formulating or proving mathematical claims or proofs, providing critical ingredients for proving mathematical claims, assisting in the writing of proofs, proposing hypotheses, designing research methodology or experiments, cleaning or reformatting datasets, supporting qualitative or thematic data analysis, or interpreting results. Our work does not involve translation, transcription, surveys, or interviews.

We also used AI tools to format scientific figures and images (but not to generate their content), edit software code, improve the readability of the text, analyze existing literature, discover research topics, and identify research gaps and relevant literature. 

We have thoroughly reviewed all LLM-generated text and code, and adapted it when necessary. The authors take full responsibility for the final content of this work, including any text, claims, or artifacts produced with the aid of generative AI.

\bibliography{iclr2027_conference}
\bibliographystyle{iclr2027_conference}

\appendix
\newpage
\section{Extended Related Works}
\label{app:extra_related}

\textbf{Drifting models\;}
Drifting has quickly been adopted in RL: as one-step policies trained online~\citep{gao2026dbpo, zhang2026podpo} and offline~\citep{koo2026dfp}, combined with critic-driven improvement in Drift Q-Learning~\citep{houssaini2026driftql}, and as fast action-conditioned world models~\citep{lu2026driftworld}.
To our knowledge, drifting has so far only been applied to targets that are given by data, like actions or next states. 
We develop a recursive temporal-difference variant for occupancy models.

\textbf{Successor measures\;} Successor measures are often learned either through representations and density ratios~\citep{touati2021, agarwal2025, bagatella2025, eysenbach2020, eysenbach2022, wiltzer2024}, or as a \emph{generative} model from which future states are sampled: $\gamma$-models~\citep{janner2020, thakoor2022}, Bellman diffusion~\citep{schramm2024}, TD-Flow~\citep{farebrother2025, farebrother2026}, flow successor representations~\citep{shi2026}, and InFOM~\citep{zheng2025}.
This work belongs to the second family and is closest to InFOM, which provides the starting pipeline that we decompose, update, and streamline.

\textbf{Offline policy extraction\;}
Offline policy extraction must stay close to the data~\citep{levine2020}, e.g.\ via conservative values~\citep{kumar2020cql}, in-sample expectile regression~\citep{kostrikov2021}, advantage-weighted regression~\citep{peng2019}, or behavior-regularized deterministic policy gradients~\citep{silver2014} such as TD3+BC~\citep{fujimoto2021} and ReBRAC~\citep{tarasov2023rebrac}, while gradient-based extraction tends to outperform weighted regression~\citep{park2024bottleneck}. 
Generative occupancy models have so far distilled an explicit critic~\citep{zheng2025}, since gradients through iterative samplers are expensive and poorly conditioned~\citep{suh2022}. 
We study whether a one-step occupancy model enables direct, critic-free extraction.

\section{Additional Results}

\begin{table}[ht]
    \centering
    \scriptsize
\begin{tabular}{l r@{\,{\scriptsize$\pm$}\,}l r@{\,{\scriptsize$\pm$}\,}l r@{\,{\scriptsize$\pm$}\,}l r@{\,{\scriptsize$\pm$}\,}l r@{\,{\scriptsize$\pm$}\,}l r@{\,{\scriptsize$\pm$}\,}l r@{\,{\scriptsize$\pm$}\,}l r@{\,{\scriptsize$\pm$}\,}l}
\toprule
Task & \multicolumn{2}{c}{BC} & \multicolumn{2}{c}{\textcolor{gray}{FB}} & \multicolumn{2}{c}{InFOM} & \multicolumn{2}{c}{FOM} & \multicolumn{2}{c}{InDOM} & \multicolumn{2}{c}{DOM} & \multicolumn{2}{c}{InDOM-Direct} & \multicolumn{2}{c}{DOM-Direct} \\
\midrule
\multicolumn{17}{l}{\textbf{\texttt{cheetah}}} \\
Walk & $212$ & {\tiny$9$} & \textcolor{gray}{$892$} & \textcolor{gray}{{\tiny$32$}} & $\mathbf{498}$ & {\tiny$14$} & $478$ & {\tiny$23$} & $432$ & {\tiny$25$} & $395$ & {\tiny$15$} & $438$ & {\tiny$26$} & $362$ & {\tiny$21$} \\
Walk Backward & $256$ & {\tiny$6$} & \textcolor{gray}{$913$} & \textcolor{gray}{{\tiny$35$}} & $540$ & {\tiny$45$} & $527$ & {\tiny$50$} & $609$ & {\tiny$30$} & $637$ & {\tiny$38$} & $\mathbf{702}$ & {\tiny$59$} & $576$ & {\tiny$64$} \\
Run & $45$ & {\tiny$3$} & \textcolor{gray}{$469$} & \textcolor{gray}{{\tiny$8$}} & $96$ & {\tiny$3$} & $\mathbf{97}$ & {\tiny$3$} & $\mathbf{97}$ & {\tiny$8$} & $62$ & {\tiny$2$} & $81$ & {\tiny$6$} & $56$ & {\tiny$2$} \\
Run Backward & $54$ & {\tiny$2$} & \textcolor{gray}{$252$} & \textcolor{gray}{{\tiny$30$}} & $84$ & {\tiny$3$} & $81$ & {\tiny$3$} & $101$ & {\tiny$12$} & $84$ & {\tiny$4$} & $\mathbf{105}$ & {\tiny$11$} & $68$ & {\tiny$4$} \\
\cmidrule{1-17}
\textit{Mean} & $142$ & {\tiny$2$} & \textcolor{gray}{$631$} & \textcolor{gray}{{\tiny$16$}} & $305$ & {\tiny$13$} & $296$ & {\tiny$15$} & $310$ & {\tiny$13$} & $295$ & {\tiny$9$} & $\mathbf{331}$ & {\tiny$14$} & $266$ & {\tiny$14$} \\
\midrule
\multicolumn{17}{l}{\textbf{\texttt{walker}}} \\
Flip & $249$ & {\tiny$3$} & \textcolor{gray}{$926$} & \textcolor{gray}{{\tiny$26$}} & $351$ & {\tiny$3$} & $394$ & {\tiny$12$} & $422$ & {\tiny$8$} & $367$ & {\tiny$6$} & $\mathbf{425}$ & {\tiny$17$} & $363$ & {\tiny$12$} \\
Run & $90$ & {\tiny$1$} & \textcolor{gray}{$367$} & \textcolor{gray}{{\tiny$12$}} & $110$ & {\tiny$2$} & $119$ & {\tiny$1$} & $119$ & {\tiny$1$} & $113$ & {\tiny$1$} & $\mathbf{125}$ & {\tiny$1$} & $112$ & {\tiny$1$} \\
Stand & $396$ & {\tiny$1$} & \textcolor{gray}{$940$} & \textcolor{gray}{{\tiny$7$}} & $601$ & {\tiny$32$} & $778$ & {\tiny$16$} & $800$ & {\tiny$8$} & $620$ & {\tiny$66$} & $\mathbf{817}$ & {\tiny$13$} & $670$ & {\tiny$50$} \\
Walk & $192$ & {\tiny$3$} & \textcolor{gray}{$886$} & \textcolor{gray}{{\tiny$29$}} & $395$ & {\tiny$16$} & $\mathbf{571}$ & {\tiny$13$} & $437$ & {\tiny$12$} & $389$ & {\tiny$8$} & $528$ & {\tiny$31$} & $413$ & {\tiny$32$} \\
\cmidrule{1-17}
\textit{Mean} & $232$ & {\tiny$2$} & \textcolor{gray}{$780$} & \textcolor{gray}{{\tiny$12$}} & $364$ & {\tiny$7$} & $466$ & {\tiny$7$} & $445$ & {\tiny$3$} & $372$ & {\tiny$15$} & $\mathbf{474}$ & {\tiny$13$} & $390$ & {\tiny$15$} \\
\midrule
\multicolumn{17}{l}{\textbf{\texttt{quadruped}}} \\
Walk & $344$ & {\tiny$3$} & \textcolor{gray}{$585$} & \textcolor{gray}{{\tiny$15$}} & $350$ & {\tiny$7$} & $307$ & {\tiny$11$} & $397$ & {\tiny$12$} & $467$ & {\tiny$12$} & $410$ & {\tiny$15$} & $\mathbf{519}$ & {\tiny$22$} \\
Jump & $616$ & {\tiny$3$} & \textcolor{gray}{$630$} & \textcolor{gray}{{\tiny$25$}} & $561$ & {\tiny$27$} & $481$ & {\tiny$15$} & $702$ & {\tiny$3$} & $754$ & {\tiny$5$} & $710$ & {\tiny$4$} & $\mathbf{783}$ & {\tiny$8$} \\
Run & $347$ & {\tiny$2$} & \textcolor{gray}{$416$} & \textcolor{gray}{{\tiny$14$}} & $339$ & {\tiny$10$} & $367$ & {\tiny$8$} & $414$ & {\tiny$2$} & $455$ & {\tiny$8$} & $421$ & {\tiny$2$} & $\mathbf{479}$ & {\tiny$2$} \\
Stand & $696$ & {\tiny$4$} & \textcolor{gray}{$775$} & \textcolor{gray}{{\tiny$32$}} & $584$ & {\tiny$112$} & $619$ & {\tiny$21$} & $827$ & {\tiny$3$} & $945$ & {\tiny$4$} & $829$ & {\tiny$6$} & $\mathbf{949}$ & {\tiny$3$} \\
\cmidrule{1-17}
\textit{Mean} & $501$ & {\tiny$3$} & \textcolor{gray}{$601$} & \textcolor{gray}{{\tiny$18$}} & $458$ & {\tiny$37$} & $443$ & {\tiny$9$} & $585$ & {\tiny$4$} & $655$ & {\tiny$4$} & $592$ & {\tiny$6$} & $\mathbf{682}$ & {\tiny$6$} \\
\midrule
\multicolumn{17}{l}{\textbf{\texttt{jaco}}} \\
Reach Top Left & $2$ & {\tiny$0$} & \textcolor{gray}{$1$} & \textcolor{gray}{{\tiny$0$}} & $177$ & {\tiny$4$} & $164$ & {\tiny$4$} & $153$ & {\tiny$6$} & $110$ & {\tiny$6$} & $\mathbf{196}$ & {\tiny$6$} & $156$ & {\tiny$9$} \\
Reach Top Right & $4$ & {\tiny$1$} & \textcolor{gray}{$5$} & \textcolor{gray}{{\tiny$4$}} & $157$ & {\tiny$5$} & $137$ & {\tiny$5$} & $138$ & {\tiny$9$} & $105$ & {\tiny$4$} & $\mathbf{185}$ & {\tiny$8$} & $139$ & {\tiny$4$} \\
Reach Bottom Left & $2$ & {\tiny$1$} & \textcolor{gray}{$1$} & \textcolor{gray}{{\tiny$1$}} & $190$ & {\tiny$2$} & $161$ & {\tiny$7$} & $168$ & {\tiny$6$} & $108$ & {\tiny$4$} & $\mathbf{211}$ & {\tiny$4$} & $184$ & {\tiny$5$} \\
Reach Bottom Right & $4$ & {\tiny$1$} & \textcolor{gray}{$0$} & \textcolor{gray}{{\tiny$0$}} & $176$ & {\tiny$4$} & $149$ & {\tiny$6$} & $146$ & {\tiny$2$} & $105$ & {\tiny$4$} & $\mathbf{210}$ & {\tiny$3$} & $165$ & {\tiny$8$} \\
\cmidrule{1-17}
\textit{Mean} & $3$ & {\tiny$1$} & \textcolor{gray}{$2$} & \textcolor{gray}{{\tiny$1$}} & $175$ & {\tiny$3$} & $153$ & {\tiny$2$} & $151$ & {\tiny$4$} & $107$ & {\tiny$3$} & $\mathbf{201}$ & {\tiny$5$} & $161$ & {\tiny$3$} \\
\bottomrule
\end{tabular}

    \caption{\textbf{Per-task results, all variants.} Undiscounted episode return per task (mean $\pm$ s.e.m.\ over random seeds), grouped by domain with fixed configuration per domain and method; \textit{Mean} rows give the four-task domain average. \textcolor{gray}{FB} numbers are rerun using the codebase from \citet{bagatella2025}. InFOM and FOM numbers are rerun using the codebase from \citet{zheng2025}. Bold marks the best method per row.}
    \label{tab:task_results}
\end{table}

\newpage
\section{Hyperparameters and Tuning Protocol}
\label{app:hyperparameters}

We follow InFOM's tuning protocol and inherit the following design choices. From Appendix D.4 of \citet{zheng2025}, we use $\tau = 0.9$ for the expectile loss in \Cref{eq:expectile_critic}, and the number of flow/drift samples ${N = 16}$ for the Monte Carlo $Q$ estimation. 
We do not re-tune InFOM's parameters since they belong to the pipeline InFOM already tuned per environment, and re-searching them would make our comparison to InFOM a comparison between two tuning budgets. 

We tune parameters and design choices that the drift model adds or changes: the discount factor $\gamma \in \{0.98, 0.99\}$, the kernel temperature $\kappa \in \{0.1, 0.2, (0.05, 0.1, 0.2)\}$, and the KL weight $\lambda_{\mathrm{KL}} \in \{0.1, 1, 10, 100\}$ on the intention encoder. The learning rate and target update rate are set to more conservative values (both $10^{-4}$) since we focused on avoiding divergence rather than convergence speed. The conservative choices on learning rate and target update rate also forced an increase in the number of update steps necessary compared to InFOM. For the occupancy flow and intention encoder, InFOM used $1$M steps for pre-training and $500$K update steps during policy extraction, while we used $1.5$M steps for cheetah and $2$M steps for the other domains for pre-training, and kept the occupancy model frozen during policy extraction. 

For our pipeline we tune the discount factor $\gamma \in \{0.98, 0.99\}$ based on the \emph{unconditional} pipeline where no KL term exists per domain, and used the optimized discounts for the intention-conditioned runs for the domain task pairs: \texttt{cheetah-walk}, \texttt{quadruped-walk}, \texttt{walker-walk}, and \texttt{jaco-reach-top-left}, where $\lambda_{\mathrm{KL}} \in \{0.1, 1, 10, 100\}$ is then swept on top of them. For a fair comparison with InFOM, we also tuned the discount factor per domain, keeping the rest of the hyperparameters as reported in \citet{zheng2025}. 

\begin{table}[]
    \centering
    \scriptsize
    \begin{tabular}{l | ll}
        \toprule
        Hyperparameter & (In)DOM & (In)FOM \\
        \midrule
        learning rate                       & $10^{-4}$                      & $3 \times 10^{-4}$ \\
        optimizer                           & Adam                           & Adam \\
        pre-training steps                   & $1.5$M (\texttt{cheetah}), $2$M (else)  & $1$M \\
        policy extraction / finetuning steps & $500$K                        & $500$K \\
        batch size                          & $256$                            & $256$ \\
        target network update rate          & $10^{-4}$                      & $5 \times 10^{-4}$ \\
        double-Q aggregation                & min                             & min \\
        policy update frequency             & $0.25$                       & $0.25$ \\
        IQL expectile                        & $0.9$                            & $0.9$ \\
        \# bootrstrap samples $N^+$         & $100$                             & -- \\
        \bottomrule
    \end{tabular}
    \caption{\textbf{Common hyperparameters for our method and baselines}. For direct policy extraction via (In)DOM, we inherit all relevant hyperparameters except the IQL expectile.}
    \label{tab:common_hyp}
\end{table}

\begin{table}[ht]
    \centering
    \scriptsize
\begin{tabular}{l|l c c c c c c}
    \toprule
    Environment & Parameter & InFOM & FOM & InDOM & DOM & InDOM-Direct & DOM-Direct \\
    \midrule
    \texttt{cheetah} & $\gamma$  & 0.98 & 0.98 & 0.99 & 0.99 & 0.99 & 0.99 \\
        & $\lambda_\mathrm{KL}$ & 0.05 & -- & 10 & -- & 10 & -- \\
        & $\alpha$  & 0.3 & 0.3 & 0.3 & 0.3 & 0.3 & 0.3 \\
        & $\kappa$  & -- & -- & 0.2 & $(0.05, 0.1, 0.2)$ & 0.1 & 0.1 \\
        & $M$       & -- & -- & -- & -- & 10 & -- \\
        & dim$(z)$  & 128 & -- & 128 & -- & 128 & -- \\
    \midrule
    \texttt{quadruped} & $\gamma$  & 0.98 & 0.98 & 0.98 & 0.98 & 0.98 & 0.98 \\
          & $\lambda_\mathrm{KL}$ & 0.005 & -- & 100 & -- & 100 & -- \\
          & $\alpha$  & 0.3 & 0.3 & 0.3 & 0.3 & 0.3 & 0.3 \\
          & $\kappa$  & -- & -- & 0.1 & 0.1 & 0.1 & $(0.05, 0.1, 0.2)$ \\
          & $M$       & -- & -- & -- & -- & 1 & -- \\
          & dim$(z)$  & 512 & -- & 512 & -- & 512 & -- \\
    \midrule
    \texttt{walker} & $\gamma$  & 0.98 & 0.98 & 0.98 & 0.98 & 0.98 & 0.98 \\
           & $\lambda_\mathrm{KL}$ & 0.1 & -- & 100 & -- & 100 & -- \\
           & $\alpha$  & 0.3 & 0.3 & 0.3 & 0.3 & 0.3 & 0.3 \\
           & $\kappa$  & -- & -- & 0.1 & 0.2 & 0.2 & 0.2 \\
           & $M$       & -- & -- & -- & -- & 10 & -- \\
           & dim$(z)$  & 512 & -- & 128 & -- & 128 & -- \\
    \midrule
    \texttt{jaco} & $\gamma$  & 0.99 & 0.99 & 0.99 & 0.99 & 0.98 & 0.98 \\
         & $\lambda_\mathrm{KL}$ & 0.2 & -- & 100 & -- & 100 & -- \\
         & $\alpha$  & 0.1 & 0.1 & 0.3 & 0.3 & 0.3 & 0.3 \\
         & $\kappa$  & -- & -- & 0.2 & 0.2 & 0.1 & 0.1 \\
         & $M$       & -- & -- & -- & -- & 10 & -- \\
         & dim$(z)$  & 512 & -- & 128 & -- & 128 & -- \\
    \bottomrule
    \end{tabular}
    \caption{\textbf{Domain-specific hyperparameters for our method and the baselines}. }
    \label{tab:domain_hyp}
\end{table}

\subsection{Evaluation Protocol}
\label{app:evaluation_protocol}

We tune per \emph{domain}: a configuration is scored by the mean over the domain's four tasks, averaged over three pre-training seeds, and only configurations with the complete $3 \times 4$ grid are scored at all. 
For the results reported in \Cref{tab:main} and \ref{tab:task_results}, as well as for method comparisons in \Cref{fig:Q1InFOMvsInDOM}, \ref{fig:Q2_intention_ablation}, and \ref{fig:Q3_policy_extraction}, we report the best-performing configuration per method with five seeds.

\paragraph{Baselines} We report three baselines to our contribution: behavior-cloning (BC), Intention-Conditioned-Flow-Occupancy-Models (InFOM), and Forward-Backward (FB). BC represents the lower baseline and is used to show whether policy improvement provides a valuable signal to go beyond imitating the behavior. 
In the implementation, we reuse the existing pipeline for policy extraction but only optimize with respect to the BC term in~\Cref{eq:ddpg_bc}, also for $500$K steps.  
Since our contributions sit on the shoulders of InFOM, we rerun their pipeline with the reported best configurations for our five seeds. 
FB is not a direct fit to our setup, yet a very powerful one in zero-shot RL. Its motivation is aligned with ours: pre-training from unlabeled data and later performing fast adaptation (in their case, zero-shot) to a revealed reward. 
The reported results in \Cref{tab:task_results,tab:main} are obtained by running the codebase of TD-JEPA \citep{bagatella2025} and tuning over the orthonormalization coefficient $\{0.01, 0.1, 1\}$. We further extended the set of domains TD-JEPA supports to \texttt{jaco}, and tested the same orthonormalization grid with an additional sweep over the discount factor $\gamma \in \{0.98, 0.99\}$. 

\section{Implementation Details}
\label{app:impl_details}

We further provide details on the practical implementation of pre-training (see \Cref{alg:pre-training}) and policy extraction based on drifting occupancies via critic-based and direct gradient estimation (see \Cref{alg:(in)dom_critic} and \ref{alg:(in)dom_direct} )

\textbf{Sample based estimate of the TD-Drift field $V_{q_{\bar\theta}^\pi}^+$\;} 
To apply the TD-Drift objective derived in \Cref{eq:td_drift} in practice, we convert the attraction and repulsion forces into Monte Carlo estimators: $\hat V^{+}_{q_{\bar\theta}^\pi}$ and $\hat V^{-}_{q}$. For the attraction, we make the following simplifications: (i) $p(\cdot \mid s, a)$ is unknown; hence we estimate it via single samples from the dataset $(s, a, s') \sim \mathcal{D}$; (ii) similar to InFOM, we fit a SARSA-variant of the algorithm in which the next action $a'$ is likewise obtained from the dataset, sampled from the data-generating policy, i.e., $(s, a, s', a') \sim \mathcal{D}$; (iii) each transition provides exactly one next state $s'$, whereas the bootstrap term can be sampled arbitrarily often. We draw $N^+$ bootstrap samples $\bar s^+_j = f_{\bar\theta}(\epsilon_j \mid s', a')$ and re-weight the forces of the positive field directly:
\begin{equation}
    \hat V^{+}_{q_{\bar\theta}^\pi}(x) = \frac{(1-\gamma)\, k(x, s')\,(s' - x) + \frac{\gamma}{N^+}\sum_{j=1}^{N^+} k(x, \bar s^+_j)\,(\bar s^+_j - x)}{(1-\gamma)\, k(x, s') + \frac{\gamma}{N^+}\sum_{j=1}^{N^+} k(x, \bar s^+_j)}.
    \label{eq:weighted_attraction}
\end{equation}
In the case of a deterministic transition kernel, \Cref{eq:weighted_attraction} is a consistent estimator of \Cref{eq:td_attraction} as $N^+ \to \infty$. In contrast, a stochastic transition kernel requires an additional aggregation over next-state samples, as already indicated by \Cref{eq:td_attraction}. 
The repulsion field $\hat V^-_q$ however, as in \Cref{eq:drift_field}, is estimated from $G$ independently generated Monte Carlo samples $x_i = f^\pi_\theta(\epsilon_i \mid s, a)$.

\textbf{Final activation and observation normalization\;}
In contrast to flow-occupancies as in InFOM \citep{zheng2025}, a drift generator $f_\theta$ directly predicts in data space. To regularize bootstrap samples from the target model, flow-based samplers make a hard projection into a box-constrained manifold of the marginal state space from the dataset. Since we designed drifting occupancies with direct differentiation in mind, this non-differentiable projection is not an option since it invalidates gradients. To still satisfy the box constraint while retaining differentiability of generated samples, we use \texttt{tanh} as the final activation function. 

\texttt{Tanh} as the final activation function pairs naturally with max/min normalization of the dataset, mapping the state space into the hypercube $[-1, 1]^d$, where $d$ is the dimensionality of the state space. Although this normalization provides the advantage of narrowing down the search for feasible kernel length scales, it is also a stronger assumption than the mean/standard-deviation normalization used by the flow baselines, since a single outlying transition compresses the whole coordinate. In our implementation, the normalization statistics are computed once, on the pre-training dataset, and shared between pre-training and policy extraction, so the frozen occupancy model remains valid across both phases. All drift results reported in this paper use this normalization.

\begin{wrapfigure}{r}{2.2in}
\centering
\vspace{-2mm}
\resizebox{2.2in}{!}{%
\begin{tikzpicture}[
  font=\footnotesize,
  block/.style={
    draw, rounded corners, thick,
    minimum width=3.0cm, minimum height=1.3cm,
    align=center, fill=blue!5
  },
  noise/.style={
    draw, circle, thick, minimum size=0.8cm, fill=gray!10
  },
  attract/.style={
  -{Stealth[length=3mm]}, thick, red!70!black,
  decorate, decoration={snake, amplitude=0.35mm, segment length=2.2mm}
  },
  ema/.style={
    -{Stealth[length=3mm]}, thick, blue!60!black
  },
  feed/.style={
    -{Stealth[length=2mm]}, thick, gray!40!black
  },
  step/.style={
    font=\scriptsize\bfseries, circle, draw, inner sep=1pt,
    minimum size=4mm, fill=white
  }
]
  
\node[block] (pred) {prediction model $f_\theta$};
\node[block, below=2.0cm of pred] (targ) {target model $f_{\bar\theta}$};

\node[noise, left=2.6cm of $(pred)!0.5!(targ)$] (eps) {$\epsilon$};
\draw[feed] (eps) -- (pred.west);
\draw[feed] (eps) -- (targ.west);
\node[left=0.05cm of eps, align=right, font=\scriptsize]
  {$\mathcal{N}(0, I) \sim $};

\coordinate (predSouthR) at ($(pred.south)+(0.45,0)$);
\coordinate (predSouthL) at ($(pred.south)+(-0.45,0)$);
\coordinate (targNorthR) at ($(targ.north)+(0.45,0)$);
\coordinate (targNorthL) at ($(targ.north)+(-0.45,0)$);

\draw[attract] (targNorthR) to[bend right=35]
  node[right, midway, xshift=0.05cm, align=left, red!70!black]
    {(iii) \\ coupled- \\ attraction}
  (predSouthR);

\draw[ema] (predSouthL) to[bend right=35]
  node[left, midway, xshift=-0.05cm, align=right, blue!60!black]
    {(iv) \\ EMA}
  (targNorthL);

\end{tikzpicture}}
\vspace{-5mm}
\caption{\textbf{Positive feedback loop from noise coupling} (iii) shared $\epsilon$ biases the kernel towards self-attraction; (iv) the EMA update
feeds this bias into the target, closing the loop.}
\label{fig:noise-feedback}
\vspace{-4mm}
\end{wrapfigure}
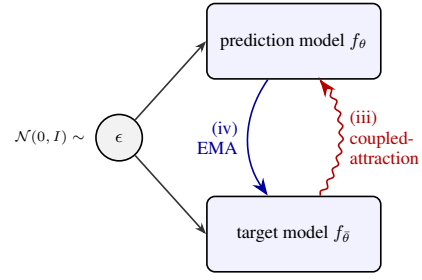

\textbf{Decoupling Noise\;} We tested the noise-coupling scheme from TD-Flow~\citep{farebrother2025}, which reuses the \emph{same} noise $\epsilon \sim \mathcal{N}(0, \text{I})$ for the current prediction $x = f_\theta^\pi(\epsilon \mid s, a)$ and the bootstrap sample $\bar s^+ = f_{\bar\theta}^\pi(\epsilon \mid s', a')$. 
We find that this coupling destabilizes training via a positive feedback loop (see Figure~\ref{fig:noise-feedback}): (i) prediction and target networks start identical since $\theta = \bar\theta$; (ii) consecutive state-action pairs within a trajectory are close $(s, a) \approx (s', a')$, so $q_\theta^\pi(\cdot \mid s,a) \overset{D}{\approx} q_{\bar\theta}^\pi(\cdot \mid s',a')$; (iii) coupling $\epsilon$ on top of this injects a bias into the kernel matrix, dominantly attracting $f_\theta^\pi(\epsilon \mid s, a)$ towards $f_{\bar\theta}^\pi(\epsilon \mid s', a')$; (iv) since $\bar\theta$ is an EMA of $\theta$, this bias feeds back into the target, closing the loop and destabilizing TD learning. Throughout this work, we therefore use i.i.d.\ $\epsilon, \bar \epsilon \sim \mathcal{N}(0, \text{I})$ for $f_\theta^\pi$ and $f_{\bar\theta}^\pi$ for an unbiased estimate of the bootstrap distribution. 

\textbf{Policy extraction\;}
In all methods, we extract a parameterized policy $\pi_\omega$ via a
behavior-regularized DDPG-BC objective~\citep{fujimoto2021}, following the action gradient $\nabla_a \tilde{Q}(s,a)$:
\begin{equation}
    \mathcal{L}_\pi(\omega) = \mathbb{E}_{(s, a) \sim \mathcal{D}_r,\, \hat a \sim \pi_\omega(\cdot \mid s)}\big[-\tilde{Q}(s, \hat a) - \alpha \log \pi_\omega(a \mid s)\big].
    \label{eq:ddpg_bc}
\end{equation}
Here, $\tilde{Q}$ can be either a Monte Carlo value estimate $\hat{Q}$ (see~\Cref{eq:q_mc}) or a distilled critic $Q_\psi$ (see~\Cref{eq:expectile_critic}). We adopt DDPG-BC because \citet{park2024bottleneck} have shown that it outperforms other behavior-regularized approaches such as weighted-regression alternatives~\citep{peng2019}.

\newcommand{\algcond}[1]{{\color{teal}#1}}
\newcommand{\alguncond}[1]{{\color{orange}#1}}
\newcommand{\algcritic}[1]{{\color{orange!80!black}#1}}
\newcommand{\algdirect}[1]{{\color{violet}#1}}

\clearpage
\newpage

\begin{algorithm}[H]
\begin{algorithmic}[1]
\caption{\textbf{InDOM / DOM: pre-training.} For intention-\algcond{conditioned} or -\alguncond{unconditioned} occupancies}
\label{alg:pre-training}
\Require \# pushforward samples $G$, \# bootstrap samples $N^+$
\State Warmup $f_\theta$ with static target $s \sim \mathcal{D}$
\State $\bar \theta \leftarrow \theta$
\While{pre-training}
\State Sample batch $\{(s, a, s', a')\} \sim \mathcal{D}$
\State Sample noise $\{\epsilon_i\}_{i=1}^G, \{\bar \epsilon_j\}_{j=1}^{N^+} \sim \mathcal{N}(0, \text{I})$
\item[]
\LeftComment {\textcolor{gray}{Infer Policy Intention}}
\State \algcond{$z \sim p_e(\cdot \mid s', a')$}
\State \algcond{$\mathrm{KL} = D_{\mathrm{KL}}(p_e(z \mid s', a') \,\|\, \mathcal{N}(0, \text{I}))$} 
\State \alguncond{($z \gets 0$, $\mathrm{KL} \gets 0$)}
\item[]
\LeftComment {\textcolor{gray}{pushforward, target-network bootstrap}}
\State $x_i = f_{\theta}^\pi(\epsilon_i \mid s, a, z)$ 
\State $\ \bar s^+_j = \mathrm{sg}\big(f_{\bar\theta}^\pi(\bar\epsilon_j \mid s', a', z)\big)$ \
\State $\hat V^+_{q_{\bar\theta}^\pi}(x_i)$ from \Cref{eq:weighted_attraction} 
\State $\mathcal{L}_{\mathrm{Drift}}^\pi = \big\|x_i - \mathrm{sg}(x_i + \hat V^+_{q_{\bar\theta}^\pi} (x_i) - \hat V^-_q(x_i))\big\|^2$
\item[]
\State Update $\theta$ (\algcond{and $e$}) to minimize $\mathcal{L}_{\mathrm{Drift}}^\pi + \algcond{\lambda_{\mathrm{KL}}\,\mathrm{KL}}$
\State update $\omega$ to minimize BC loss $-\log \pi_\omega(a \mid s)$
\item[]
\LeftComment {\textcolor{gray}{EMA target update}}
\State $\bar \theta \leftarrow \eta \theta + (1 - \eta) \bar \theta$ 
\EndWhile
\State Freeze $\theta$ (\algcond{and $e$})
\end{algorithmic}
\end{algorithm}

\begin{minipage}{0.49\textwidth}
\begin{algorithm}[H]
    \small
    \centering
    \caption{\textbf{Critic-Based Policy Extraction} for intention-\algcond{conditioned} or -\alguncond{unconditioned} occupancies}
    \label{alg:(in)dom_critic}    
    \begin{algorithmic}[1]
        \Require \# MC-samples $N$
        \While{policy extraction}
        \State Sample reward batch $\{(s, a, r)\} \sim \mathcal{D}_r$
        \State Sample noise $\{\epsilon_i\}_{i=1}^N \sim \mathcal{N}(0, \text{I})$
        \item[]
        \State update $\phi$ to minimize $\mathbb{E}\big[\|r_\phi(s) - r\|^2\big]$
        \item[]
        \LeftComment {\textcolor{gray}{Prepare GPI}}
        \State \algcond{sample latent $z \sim \mathcal{N}(0, \text{I})$}
        \State \alguncond{($z \gets 0$)}
        \item[]
        \LeftComment {\textcolor{gray}{Make MC value estimate}}
        \State $\hat s^+_{i} = f_\theta^\pi(\epsilon_i \mid s, a, z) $
        \State $\hat Q_z(s, a) = \frac{1}{(1 - \gamma) N} \sum_i r_\phi(\hat s^+_{i})$
        \item[]
        \LeftComment {\textcolor{gray}{GPI/PI over intentions}}
        \State \algcond{$\psi \leftarrow \psi - \nabla_\psi \mathbb{E}\big[\ell_2^\tau(\hat Q_z(s, a) - Q_\psi(s, a))\big]$}
        \State \alguncond{$\psi \leftarrow \psi - \nabla_\psi \mathbb{E}\big[\|\hat Q_z(s, a) - Q_\psi(s, a)\|^2\big]$}
        \item[]
        \LeftComment{\textcolor{gray}{DDPG-BC actor update}}
        \State $\hat a \sim \pi_\omega(\cdot \mid s)$
        \State $\omega \leftarrow \omega  + \nabla_\omega (\mathbb{E}[Q_\psi(s, \hat a) + \alpha \log \pi_\omega(a \mid s)])$ 
        \EndWhile
    \end{algorithmic}
\end{algorithm}
\end{minipage}
\hfill
\begin{minipage}{0.49\textwidth}
\begin{algorithm}[H]
    \small 
    \centering
    \caption{\textbf{Direct Policy Extraction} for intention-\algcond{conditioned} or -\alguncond{unconditioned} occupancies}
    \label{alg:(in)dom_direct}
    \begin{algorithmic}[1]
        \Require \# MC-samples $N$, \# latent candidates $M$
        \While{policy extraction}
        \State Sample reward batch $\{(s, a, r)\} \sim \mathcal{D}_r$
        \State Sample noise $\{\epsilon_i\}_{i=1}^N \sim \mathcal{N}(0, \text{I})$
        \item[]
        \State update $\phi$ to minimize $\mathbb{E}\big[\|r_\phi(s) - r\|^2\big]$
        \item[]
        \LeftComment {\textcolor{gray}{Prepare GPI}}
        \State \algcond{sample $M$ latents $\{z_m \sim \mathcal{N}(0, \text{I})\}_{m=1}^M$}
        \State \alguncond{($z_m \gets 0$)}
        \item[]
        \LeftComment {\textcolor{gray}{Make MC value estimate}}
        \State $\hat a \sim \pi_\omega(\cdot \mid s)$
        \State $\hat s^+_{i, m} = f_\theta^\pi(\epsilon_i \mid s, \hat a, z_m) $
        \State $\hat Q_{z_m}(s, \hat a) = \frac{1}{(1 - \gamma) N} \sum_i r_\phi(\hat s^+_{i, m})$
        \item[]
        \LeftComment {\textcolor{gray}{GPI/PI over intentions}}
        \State \algcond{$\hat Q(s, \hat a) = \max_m \hat Q_{z_m}(s, \hat a)$} 
        \State \alguncond{$\hat Q(s, \hat a) = \hat Q_0(s, \hat a)$}
        \item[]
        \LeftComment{\textcolor{gray}{DDPG-BC actor update}}
        \State $\omega \leftarrow \omega +\nabla_\omega (\mathbb{E}[\hat Q(s, \hat a) + \alpha \log \pi_\omega(a \mid s)])$ 
        \EndWhile
    \end{algorithmic}
\end{algorithm}
\end{minipage}

\clearpage
\newpage

\section{Ablations}

\begin{wrapfigure}[16]{r}{2.75in}
    \centering
    \vspace{-4mm}
    \includegraphics[width=\linewidth]{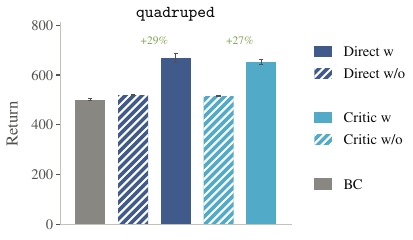}
    \vspace{-7mm}
    \caption{\textbf{Effect of warming up initial spread}. Policy return on \texttt{quadruped} for DOM and DOM-Direct with (w) and without (w/o) warm-up phase to generate an initial spread. }
    \vspace{-4mm}
    \label{fig:discount_warmup_bars}
\end{wrapfigure}

\subsection{An initial spread makes TD-Drift work}
\label{app:warmup}
The bootstrapped objective guides the model towards the target distribution through the kernel similarity between its own bootstrap samples and the observed next states (see \Cref{eq:weighted_attraction}).
If the initial distribution is concentrated, the kernel similarity is uniformly high towards the bootstrap samples, since those are very close. If the learning problem were set up with a high enough discount factor $\gamma \to 1$, the supervision signal for the next state approaches $0$, causing the prediction model to chase after its bootstrap estimates, a positive feedback loop similar to that in \Cref{fig:noise-feedback}. 
To avoid this, we propose an initial warm-up pre-training phase ($10$K steps) to spread the initial drifting distribution -- for instance, to approximate the coverage of the dataset.
This shortens the expected distance to the next state, raising the expected kernel similarity to the next state supervision signal into its informative range.
We find this initial spread to be crucial for the TD-Drift objective to work. To ablate this, we evaluate DOM and DOM-Direct on the \texttt{quadruped} domain over four tasks using drifting models with and without this initial warm-up phase.
\Cref{fig:discount_warmup_bars} shows that all methods using an occupancy without (w/o) warm-up collapse back to the BC baseline performance.

\begin{wrapfigure}{r}{2.75in}
    \centering
    \vspace{-4mm}
    \includegraphics[width=\linewidth]{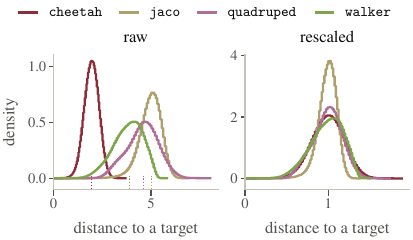}
    \vspace{-7mm}
    \caption{\textbf{Rescaling kernel distances per dataset}. Two histograms of inter-sample distances. The left one is based on min-max normalization per data domain. Right one: each distance is divided by the median pairwise distance (dashed line in left histogram)}
    \label{fig:distances}
    \vspace{-4mm}
\end{wrapfigure}

\subsection{Median temperature rescaling}
\label{app:temp_rescaling}
Drifting is, at its core, a kernel method, and inherits the problem of choosing a kernel bandwidth. 
\citet{deng2026} mitigate this by aggregating the drift field over an array of temperatures, offering potential robustness. Other works like \citet{houssaini2026driftql} have shown drifting to work well also with a single temperature, but the search space across different datasets and geometries is still vast. 
Our TD-Drift implementation in \Cref{alg:pre-training} already normalizes the state space to a $[-1, 1]^d$ hypercube, which fixes the \emph{range} of the sample coordinates but not the \emph{scale} of pairwise distances between samples, since this scale still depends on the dimensionality and geometry of each domain.

After priming the model to approximate the state space coverage sufficiently well through the warm-up phase described in \Cref{app:warmup}, we are, at the start of TD-learning, left with a fixed but a priori unknown distribution of inter-sample distances.
At the start of occupancy-learning through the TD-objective, we are therefore interested in how well a kernel of a given temperature can pick up the signal from the next-state supervision. 

Since this informative range depends on the dataset's own distance scale, which we do not know ahead of time, we rescale the temperatures by the median expected inter-sample distance of the dataset
(see \Cref{fig:distances}), which unifies the distance scale across domains and makes a single default temperature(-field) choice meaningful.
Rescaling is used throughout the paper. Despite that, the chosen temperature field $(0.05, 0.1, 0.2)$ is not a clear win, since it is reported to be best in only two out of 16 domain $\times$ task pairs; it 
usually is not far from the best tested temperature and therefore represents a reasonable default.

\end{document}